\documentclass[conference]{IEEEtran}
\IEEEoverridecommandlockouts
\usepackage{cite}
\usepackage{amsmath,amssymb,amsfonts}
\usepackage{bbm}
\usepackage{algorithmic}
\usepackage{graphicx}
\usepackage{textcomp}
\usepackage{xcolor}
\usepackage{booktabs}
\usepackage{makecell}
\usepackage{enumerate}
\usepackage{amsthm}
\usepackage{algorithm}
\usepackage{mathtools}
\usepackage{etoolbox}
\usepackage{apptools}
\usepackage{float}
\usepackage{caption}
\usepackage{subcaption}
\usepackage{multirow}
\usepackage{balance}
\usepackage{xspace}
\usepackage{pifont}
\usepackage{url}
\usepackage{color, colortbl}
\BeforeBeginEnvironment{appendices}{\clearpage}
\newtheorem{assumption}{\textbf{Assumption}}
\newtheorem{lemma}{\textbf{Lemma}}
\newtheorem{theorem}{\textbf{Theorem}}
\theoremstyle{definition}

\newcommand{\bluenote}[1]{\textcolor{black}{#1}}

\def\ie{\textit{i.e.}\xspace}
\def\etal{\textit{et al.}\xspace}

\def\eg{\textit{e.g.}\xspace}

\DeclareMathOperator*{\argmax}{arg\,max}

\def\method{SemiSFL\xspace}

\def\BibTeX{{\rm B\kern-.05em{\sc i\kern-.025em b}\kern-.08em 
 T\kern-.1667em\lower.7ex\hbox{E}\kern-.125emX}}

\def\InvReg{global updating\xspace}
\def\Reg{cross-entity updating\xspace}
\def\locReg{cross-entity semi-supervised training\xspace}
\def\srvUp{supervised training\xspace}
\def\srvmodh{\boldsymbol{w}^{h+}}

\begin{document}

\title{SemiSFL: Split Federated Learning on Unlabeled and Non-IID Data
}

\author{
    \IEEEauthorblockN{
    Yang Xu$^{1,2}$~~Yunming Liao$^{1,2}$~~Hongli Xu$^{1,2}$~~Zhipeng Sun$^{1,2}$~~Liusheng Huang$^{1,2}$~~Chunming Qiao$^{3}$
    }
    \IEEEauthorblockA{
    $^{1}$School of Computer Science and Technology, University of Science and Technology of China\\
    $^{2}$Suzhou Institute for Advanced Research, University of Science and Technology of China\\
    $^{3}$Department of Computer Science and Engineering, University at Buffalo, the State University of New York\\
    \{xuyangcs, xuhongli, lshuang\}@ustc.edu.cn,~\{ymliao98, rodman\}@mail.ustc.edu.cn, ~qiao@buffalo.edu
    }
}

\maketitle

\thispagestyle{plain}

\begin{abstract}
Federated Learning (FL) has emerged to allow multiple clients to collaboratively train machine learning models on their private data at the network edge.
However, training and deploying large-scale models on resource-constrained devices is challenging.
Fortunately, Split Federated Learning (SFL) offers a feasible solution by alleviating the computation and/or communication burden on clients.
However, existing SFL works often assume sufficient labeled data on clients, which is usually impractical.
Besides, data non-IIDness poses another challenge to ensure efficient model training.
To our best knowledge, the above two issues have not been simultaneously addressed in SFL.
Herein, we propose a novel Semi-supervised SFL system, termed SemiSFL, which incorporates clustering regularization to perform SFL with unlabeled and non-IID client data.
Moreover, our theoretical and experimental investigations into model convergence reveal that the inconsistent training processes on labeled and unlabeled data have an influence on the effectiveness of clustering regularization.
To mitigate the training inconsistency, we develop an algorithm for dynamically adjusting the global updating frequency, so as to improve training performance.
Extensive experiments on benchmark models and datasets show that our system provides a 3.8$\times$ speed-up in training time, reduces the communication cost by about 70.3\% while reaching the target accuracy, and achieves up to 5.8\% improvement in accuracy under non-IID scenarios compared to the state-of-the-art baselines.
\end{abstract}


\section{Introduction}\label{intro}


Recently, vast amounts of data generated by mobile and Internet of Things (IoT) devices have exhibited great potential for improving the performance of various applications. 
To process these data from both privacy and economic perspectives, Federated Learning (FL) \cite{konevcny2016federated, li2019edge, xu2022adaptive} has emerged and been extensively applied in many AI applications, including intelligent medical \cite{liang2022new}, finance \cite{wen2023survey} and smart home \cite{li2020review}.
In order to capture complex patterns and relationships in massive data, large-scale models have achieved better generalization and higher accuracy in various tasks \cite{liao2023accelerating}.
However, hardware limitations of clients and network bandwidth constraints between clients and remote servers pose restrictions on typical FL.
This motivates Split Federated Learning (SFL) \cite{han2021accelerating, thapa2022splitfed, liao2024mergesfl}, a promising solution for efficiently training large-scale models. 

In SFL, a full model is divided into two submodels, \ie, a top model (close to the output layer) and a bottom model (close to the input layer), at an intermediate layer (called \textit{split layer}). 
The top model is updated on the server, while the bottom model is trained on each client.
During training, clients transmit activations (also called \textit{features}) of the split layer to the server, which further performs forward and backward propagation on the top model with the features and returns the corresponding gradients. 
For example, given a 13-layer VGG13 \cite{simonyan2014very} with size of 508 MB, when splitting the model at the 10th layer, the sizes of its bottom model and features (with batch size of 64) are about 36 MB and 2 MB, respectively.
Accordingly, model splitting contributes to reducing computation burden and communication costs for clients.
Besides, since clients have access to only the bottom models, the privacy of the model's complete structure is protected.

Despite the above benefits of SFL, two critical issues have to be addressed in SFL. 
Firstly, it is usually impractical that SFL always has access to fully labeled data for model training.
Considering that data annotation is time-consuming and probably requires domain-specific expert knowledge, it is challenging to obtain sufficient labels, which makes large amounts of valuable unlabeled data untapped.
Secondly, the clients usually belong to different individuals and/or work under different circumstances, thus data generated by different clients probably are non-Independent and Identically Distributed (non-IID).
Models trained on non-IID data across different clients usually exhibit varying divergences \cite{zhao2018federated, wang2023distribution, liao2023decentralized, gui2023sk}, resulting in performance degradation. 

To our best knowledge, the above two issues have not been addressed in SFL literature up to now.
We are encouraged to find some inspiration in previous FL works.
To make full use of unlabeled data, many FL works \cite{lin2021semifed, kim2022federated} integrate the techniques of semi-supervised learning and propose Semi-supervised FL (termed SemiFL). 
Some SemiFL works assume clients possess labeled data while the server has unlabeled data \cite{wang2022enhancing, lin2020ensemble}.
However, considering the lack of sufficient expert knowledge and/or labor, clients probably have more unlabeled data.
Herein, we focus on a more practical scenario, where labeled data are maintained by the server and unlabeled data reside on the clients \cite{albaseer2020exploiting, zhang2021improving}.
In this case, SemiFL first performs server-side supervised training on the labeled data, and then conducts client-side semi-supervised training on the unlabeled data using techniques like pseudo-labeling \cite{diao2021semifl,jeong2021federated,long2020fedsiam,zhao2023does}.
For example, Diao \etal \cite{diao2021semifl} directly distribute and employ the up-to-date global model to generate pseudo-labels for the unlabeled data, and further train the local models (replicas of the global model) on clients in parallel.
As in typical FL, the non-IID issue inevitably reduces the generalization ability of trained models in \cite{diao2021semifl}.


For the non-IID issue in SemiFL, many researchers have made efforts to improve the quality of pseudo-labels.
For example, each client in \cite{jeong2021federated} is provided with multiple helpers (\ie, models of other clients with similar predictions given the same input) to generate high-quality pseudo-labels, but at the expense of increased communication costs.
To limit the communication costs, an Exponential Moving Average (EMA) model (regarded as a \textit{teacher} model) is constructed and widely utilized to guide the updating of each client's local model (regarded as a \textit{student} model) through pseudo-labeling \cite{long2020fedsiam,zhao2023does}.
As the teacher models retain not only the generalized knowledge from historical teacher models but also the personalized knowledge from recent global models, they can help to promote the semi-supervised training on clients well. 
Besides the teacher models, some literature \cite{wang2023knowledge} employs a pre-trained model to initialize the global model and guide the aggregation of student models, so as to further improve the training performance.
However, all the existing Semi-FL arts necessitate full models for semi-supervised learning on clients, which is not applicable in SFL, since clients possess only the bottom models.

To advance SFL in utilizing unlabeled data, it is intuitive to incorporate semi-supervised learning in SFL.
In a naive Semi-supervised SFL solution, both the global model and the teacher model are split and their bottom models are allocated to the clients.
It mainly involves two alternate phases in a training round: i) server-side supervised training, where the server updates the global and teacher models with generally IID labeled data, and ii) cross-entity semi-supervised training, 
where client-side teacher/student bottom models produce the features (referred to as teacher/student features) with non-IID data.
These features are separately sent and fed to the server-side top model for pseudo-labeling and model training.
However, this solution only adopts the output logits and pseudo-labels for loss calculation, failing to adequately tackle the non-IID issue (refer to Section \ref{evaluation}).


To this end, we review the distinct properties of SFL and propose a novel Semi-supervised SFL system, termed SemiSFL, that incorporates \textit{Clustering Regularization} to perform SFL with unlabeled data.
In SemiSFL, the teacher bottom models inherit the advantages (\eg, high generalization ability) of teacher models.
Considering that the student bottom models are gradually trained on the non-IID client data as training progresses, they are more likely to overfit the client data compared to the teacher bottom models.
Thus, the student features usually suffer from non-negligible \textit{feature shift} \cite{li2021fedbn} and reduced representative ability. 
Fortunately, the clustering of teacher features provides a comprehensive view of the feature distribution, which is robust to data skewness.
As a result, the clustering results can be utilized to regularize the student bottom models, so as to improve their generalization ability, which motivates the design of clustering regularization.
Furthermore, according to our theoretical and experimental analysis in Section \ref{analysis-section}, the global updating frequency (\ie, the number of times when performing server-side supervised training in a round) has a complicated influence on model convergence, and it is quite challenging to adaptively adjust the global updating frequency to achieve satisfied training performance.
The main contributions are summarized as follows:
\begin{itemize}
    \item We develop a novel Semi-supervised SFL system, termed \method, incorporating clustering regularization, which to the best of our knowledge is the first solution to simultaneously address the issues of unlabeled and non-IID data in SFL.
    \item We theoretically and experimentally investigate the impact of global updating frequency on model convergence.
    Building on these findings, we dynamically and adaptively regulate the global updating frequency in response to expected changes in supervised
    loss to guarantee the effectiveness and efficiency of our proposed system.
    \item We conduct extensive experiments to evaluate the performance of \method. 
    The experimental results show \method provides a 3.0$\times$ speed-up in training time by reducing 70.3\% of the communication cost while reaching the target accuracy. 
    Besides, it achieves accuracy improvements up to 5.1\% under different non-IID settings compared to the state-of-the-art baselines.
\end{itemize}


\section{Background and Motivation}\label{bg}

In this section, we provide a description of Semi-supervised SFL, followed by an exploration of the underlying motivation behind the design of \method.


\subsection{Semi-supervised Split Federated Learning} \label{description-section}
Split Federated Learning (SFL) \cite{thapa2022splitfed} is a combination of Federated Learning (FL) \cite{konevcny2016federated} and Split Learning (SL) \cite{gupta2018distributed}. 
FL is designed to train ML models on distributed edge devices with privacy-preserving, while SL aims at reducing the computation cost of participants with limited resources. 
Therefore, SFL inherits the advantages of both FL and SL. 

Similar to FL, a Parameter Server (PS) and $N$ clients collaboratively train an ML model (denoted as $\boldsymbol{w}$) in SFL. 
The model is split into two submodels as $\boldsymbol{w} = (\boldsymbol{w}_s, \boldsymbol{w}_c)$. 
The bottom model $\boldsymbol{w}_c$ resides on clients while the top model $\boldsymbol{w}_s$ resides on the PS. 
In general, the PS randomly selects a subset $V_h \in V$, consisting of $N_h = |V_h|$ ($N_h \leq N$) active clients at round $h$. 
The selected clients with bottom models are instructed to send features \cite{thapa2022splitfed} and ground-truth labels $y_{i}$ ($i\in[V_h]$) to the PS. 
The PS then calculates the gradients of features for each client in parallel and sends them to clients while accumulating the gradients $\Delta \boldsymbol{w}_s$ of the top model. 
Afterward, the clients perform backpropagation to update their bottom models.
At the end of each round, the PS aggregates bottom models to obtain a global model for further training.

However, considering the lack of sufficient expert knowledge or labor on clients, it is usually practical that most or even all data on clients are unlabeled while the PS possesses some labeled data annotated by domain experts \cite{diao2021semifl,zhao2023does}.
Thus, it is appealing to develop Semi-supervised SFL,
where the complete dataset $\mathcal{D}$ consists of labeled dataset $\mathcal{D}_l$ and unlabeled dataset $\mathcal{D}_u$, and $\mathcal{D}_u \triangleq \mathcal{D}_{u, 1} \cup \mathcal{D}_{u, 2} \cup \cdots \cup \mathcal{D}_{u, N}$ ($\mathcal{D}_{u, i}$ is the dataset of client $i$). 
The loss function over a labeled data sample $(x, y)$ and the model parameter $\boldsymbol{w} \in \mathbb{R}^d$ with $d$ dimensions is defined as $\ell_{s}(x, y, \boldsymbol{w})$.
Thus, considering the supervised training on the labeled dataset $\mathcal{D}_l$, the loss function is $ \mathbb{E}_{x \in \mathcal{D}_l} \ell_s(x, y, \boldsymbol{w})$.

To leverage the vast amounts of unlabeled data on clients, recent studies have achieved promising results by enforcing model predictions on augmented data that deviate significantly from the data distribution. 
The objective is to ensure the alignment between these predictions with their corresponding pseudo-labels \cite{sohn2020fixmatch}. 
In other words, for a given unlabeled data sample $x$, the model's prediction of its weakly-augmented version is represented as a vector $q=(q_{1}, \cdots, q_{M}) \in [0, 1]^M$, where $\sum_{m=1}^M{q_{m}} = 1$, and $M$ is the number of classes. 
The pseudo-label for $x$ is then defined as $\hat{q} =\argmax_m {q_{m}}$, and is retained only if $\max_m {q_{m}}$ falls above the pre-defined confidence threshold $\tau$. 

Let $\mathcal{H}$ denote the cross-entropy loss function, which measures the discrepancy between the predicted and true labels. 
The unsupervised training loss with consistency regularization can be represented as:
\begin{equation}
     \ell_{u}(x, \boldsymbol{w}) = \mathbbm{1}(\underset{m}{\max} {(q_{m})} > \tau) \mathcal{H}(x, \hat{q}, \boldsymbol{w})
\end{equation}
Then the total training objective is expressed as:
\begin{equation}
   F(\boldsymbol{w}^*) \triangleq \underset{\boldsymbol{w} \in \mathbb{R}^d}{\min \; } \left[\mathbb{E}_{x \in \mathcal{D}_l} \ell_{s}(x, y, \boldsymbol{w}) + \mathbb{E}_{x \in \mathcal{D}_u} \ell_{u}(x, \boldsymbol{w}) \right]
\end{equation}

\begin{figure}[t]
\centering
\includegraphics[width=1.0\linewidth]{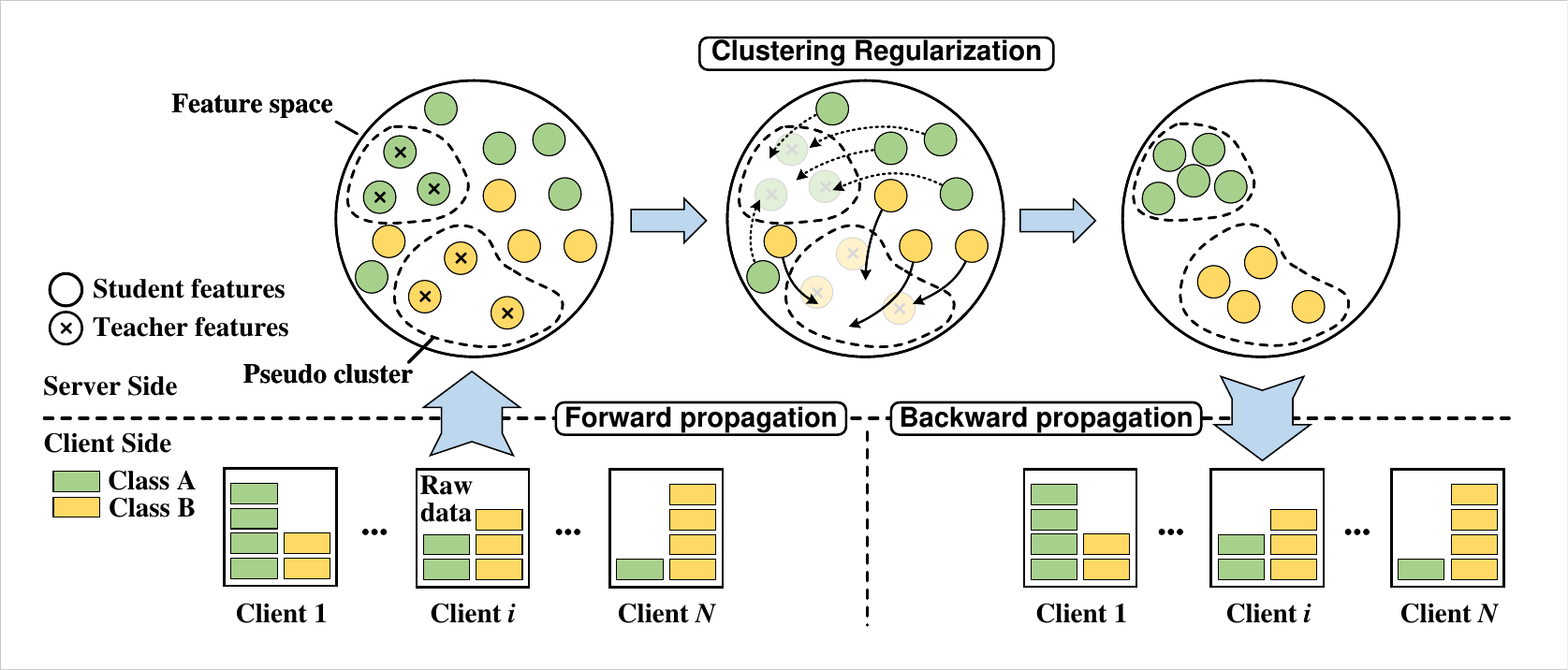}
\caption{Illustration of clustering regularization.}
\label{motiv-fig}
\vspace{-0.5cm}
\end{figure}

\subsection{Motivation for System Design} \label{motivation-section}
In SemiFL, the consistency regularization loss encounters a major hurdle when the unlabeled data is non-IID among clients, as local models tend to favor locally abundant categories while potentially disregarding minorities. 
Although each client might train an expert model, the model divergence among clients is exacerbated when the pre-defined confidence threshold ($\tau$) blocks out data samples belonging to minority categories, especially in the early stages of training. 
Consequently, such imbalance in training leads to biased features \cite{li2021model}, resulting in logits with low confidence which are more likely to be excluded from consistency regularization.
Therefore, our objective is to effectively utilize the unlabeled data by exploring inter-client knowledge while ensuring efficient communication and computation.

Fortunately, the nature of SFL provides inspiration for a solution.
In \method, pseudo-labeling is performed on the PS where the top model resides. 
As shown in Fig. \ref{motiv-fig} (left), each client transmits two types of features for loss calculation: teacher features from the teacher bottom model (used for pseudo-labeling), and student features from the local bottom model (used for computing gradients).
It is worth noting that the features collected on the PS show less skewness and higher abundance compared to those on each client. 
The fact that features are mutually accessible on the PS allows for leveraging the knowledge contained in features from multiple clients to mitigate the negative impact of non-IID data.

To this end, we propose to let the model learn more potential knowledge from features, enabling it to make confident predictions by identifying patterns and commonalities in the input.
We modify a contrastive loss \cite{lee2022contrastive} for Semi-supervised SFL and propose \textit{Clustering Regularization}.
Specifically, for efficient processing, the features are first projected into lower-dimension vectors with the help of a two-layer linear network, \ie, projection head.
Subsequently, the (low-dimension) teacher features form different clusters (called \textit{pseudo clusters}) regarding the pseudo-labels in the feature space, as illustrated in Fig. \ref{motiv-fig} (left). 
The contrastive loss then regularizes the projected student features of unlabeled samples to move towards the pseudo clusters of the same classes, as depicted in Fig. \ref{motiv-fig} (right). 
After clustering regularization, student features with the same pseudo-labels get closer to each other, which generates useful guideline information for model updating.
Owing to the computation and communication-friendly framework of SFL, the execution of clustering regularization for all participating clients is completed on the PS, relieving clients from corresponding computational and aggregation overhead.

Powered by clustering regularization, we can fully exploit the potential of unlabeled data, including the samples that are filtered by the confidence threshold and further excluded from consistency regularization. 
By leveraging the knowledge encoded in the pseudo clusters, even data samples belonging to minority classes can contribute to model training.
From another perspective, feature clustering assists and accelerates the process of consistency regularization, since well-clustered features provide data samples a higher chance to be credited with high confidence by the top model. 
Importantly, clustering regularization is not directly applied to the top model, which avoids introducing confirmation bias \cite{lee2022contrastive} and ensures that the model is always focused on the correct goal.

\section{System Design}\label{system-design}

\begin{figure}[t]
\centering
\includegraphics[width=1.0\linewidth]{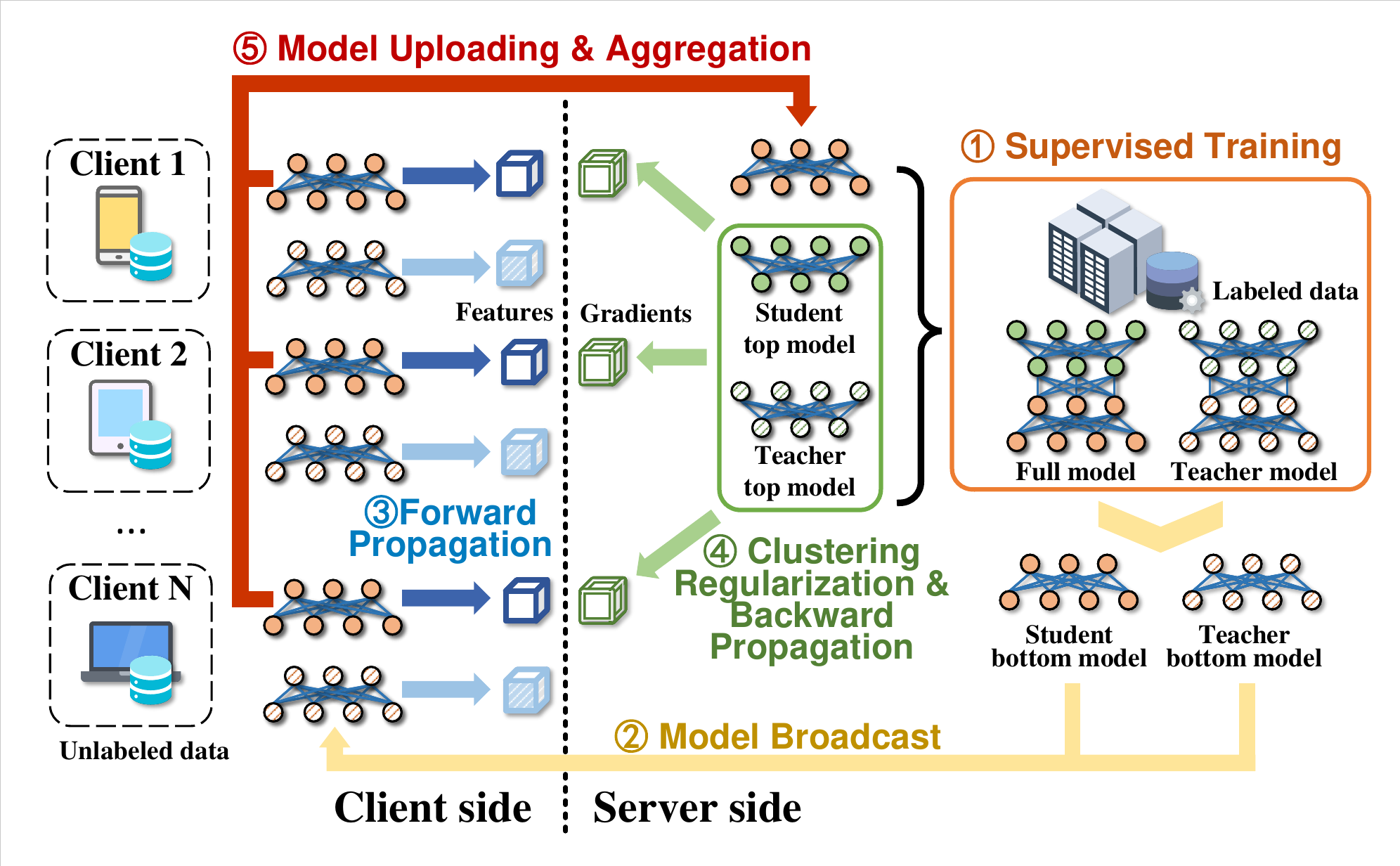}
\vspace{-0.5cm}
\caption{System workflow.}
\label{workflow-fig}
\vspace{-0.6cm}
\end{figure}

In \method, \textit{cross-entity semi-supervised training} always follows the \textit{server-side \srvUp}.
Such alternate training procedure will be conducted for $H$ rounds. 
Fig. \ref{workflow-fig} illustrates the workflow of \method, which consists of the following five processes in round $h$:

(1)-(2) \textbf{\textit{Supervised Training} and \textit{Bottom Model Broadcast}.} 
Initially, the PS trains a global model for $K_s$ iterations using labeled data.
The global model, denoted as $\boldsymbol{w}=(\boldsymbol{w}_c, \boldsymbol{w}_s)$, consists of a bottom model $\boldsymbol{w}_c$ and a top model $\boldsymbol{w}_s$.
The model is originally designed to be trained with cross-entropy loss upon model outputs and ground-truth labels. 
Besides, a supervised-contrastive loss \cite{khosla2020supervised} is also employed as the loss function, which enables the model to learn similar or dissimilar representations with labels of the same or different classes. 
To facilitate the training process, an additional projection head $\boldsymbol{w}_h$ \cite{gupta2022understanding} is juxtaposed alongside $\boldsymbol{w}_s$, receiving the output from $\boldsymbol{w}_c$.
Here, we define $g_{\boldsymbol{w}}(\cdot)$ as the feed-forward function for model $\boldsymbol{w}$. 
Let $\mathcal{B}_l \in \mathcal{D}_l$ represent any mini-batch from the labeled dataset. The loss function is defined as follows:
\begin{equation}
\label{supcon-loss}
 \mathcal{T}(x_j, \boldsymbol{w}) = \frac{-1}{|P(j)|} \sum\limits_{p \in P(j)}{\log \frac{\exp (z_j \cdot z_p / \kappa)}{ \sum\limits_{a \in A(j)}{\exp (z_j \cdot z_a / \kappa)}}}  
\end{equation}
where $z_j = g_{\boldsymbol{w}_p}(g_{\boldsymbol{w}_c}(x_j))$, $A(j)$ is the set of indexes in all reference samples except $j$, $P(j) \triangleq \{p \in A(j), y_p = y_j\}$ is the set of positive samples of image $x_j$, $\kappa \in \mathcal{R}^+$ is a scalar temperature parameter. 
In our formulation, the similar pairs consist of samples with the same label, while the dissimilar pairs consist of samples with different labels. 
The loss for supervised training is the summation over Eq. \eqref{supcon-loss} and cross-entropy loss $\mathcal{H}$:
\begin{equation}
 \ell_{s} = \mathcal{H} + \mathcal{T}
\end{equation}

During training, an EMA model $\tilde{\boldsymbol{w}}=(\tilde{\boldsymbol{w}}_c, \tilde{\boldsymbol{w}}_s, \tilde{\boldsymbol{w}}_p)$ calculated as $\tilde{\boldsymbol{w}} = \gamma \tilde{\boldsymbol{w}} + (1-\gamma) \boldsymbol{w}, \gamma \in (0, 1]$ is employed as the teacher model.
The teacher model shares the same model architecture as $\boldsymbol{w}$, with its parameters being computed as a moving average of the global models trained in previous steps.
It is utilized throughout the process to provide both pseudo labels and teacher features. 
To allow for loss calculation on abundant reference samples, we maintain a two-level memory queue $\mathcal{Q}$ on the fly that caches the most recent features generated from the teacher model. 
The gradients of the supervised-contrastive loss be only applied to the current mini-batch. 

When global training is done in round $h$, a global model $\srvmodh = \boldsymbol{w}^{h, K_s}$ is obtained. 
Part of its components including the top model $\srvmodh_s$, the projection head $\srvmodh_p$ and their teacher counterparts are kept on the server side, while the bottom model as $\srvmodh_c$ and its teacher counterpart $\tilde{\boldsymbol{w}}_c^{h +}$ are broadcast to each client. 

(3) \textbf{\textit{Forward Propagation}.} 
Each \locReg round consists of $K_u$ iterations, where $K_u$ equals the \Reg frequency. 
For client $i$, the bottom model in round $h$ is initialized as $\boldsymbol{w}_{c,i}^{h +, 1} = \boldsymbol{w}_{c}^{h +}$. Before the $k$-th iteration, the bottom model and its teacher counterpart are notated as $\boldsymbol{w}_{c,i}^{h +,k}$ and $\tilde{\boldsymbol{w}}_{c,i}^{h +,k}$, respectively.


During the $k$-th iteration, client $i$ performs \locReg on a batch size of $d_i^{h +,k}$.
Considering the feed-forward process, any data sample $x$ in a mini-batch $\mathcal{B}_{u, i} \subset \mathcal{D}_{u,i}$, $|\mathcal{B}_{u, i}| = d_i^{h +,k}$ first undergoes weak augmentation $a_w(\cdot)$, including random horizontal flipping and cropping, and strong augmentation $a_s(\cdot)$, exemplified by RandAugment \cite{cubuk2020randaugment} in our system. 
The productions are denoted as $a_w(x)$ and $a_s(x)$, respectively. 
They are then fed-forward to the bottom model $\boldsymbol{w}_{c,i}^{k,\iota}$ and teacher bottom model $\tilde{\boldsymbol{w}}_{c,i}^{k,\iota}$ in parallel, generating student features $\boldsymbol{e}_i=(e_{i,1}, \cdots, e_{i, |\mathcal{B}_{u,i}|})$ and teacher features $\tilde{\boldsymbol{e}}_i=(\tilde{e}_{i,1}, \cdots, \tilde{e}_{i, |\mathcal{B}_{u,i}|})$, where $e_{i,j} = g_{ \boldsymbol{w}_{c,i}^{h, k}}(a_s(x_{i,j}))$, $\tilde{e}_{i,j}= g_{ \tilde{\boldsymbol{w}}_{c,i}^{h, k}}(a_w(x_{i,j})), x_{i,j} \in \mathcal{B}_{u,i}$. 
The student features $\boldsymbol{e}_i$ and teacher features $\tilde{\boldsymbol{e}_i}$, each containing $d_i^{h +,k}$ samples, are then sent to the PS. 
On the PS, both the student and teacher features arriving at a synchronization barrier are then fed forward server-side models and their teacher counterparts in parallel. 

(4) \textbf{\textit{Clustering Regularization \& Backward Propagation}.} 
As stated in section \ref{description-section}, a cross-entropy loss is applied to minimize the prediction of noisy input, \ie, a strongly-augmented sample and the pseudo-label. 
Additionally, to minimize the distance of model predictions of the same class in embedding space, we extend the contrastive loss to a multi-client setting. 
Based on notations of Eq. (\ref{supcon-loss}), we define the \textit{Clustering Regularization} loss as:
\vspace{-0.3cm}
\begin{equation}
\label{cr-eq}
     \mathcal{C}(x_j, \boldsymbol{w}) = \frac{-1}{|\hat{P}(j)|} \sum\limits_{p \in \hat{P}(j)}{\log \frac{\exp (z_j \cdot \tilde{z}_{p} / \kappa)}{ \sum\limits_{a \in [\mathcal{Q}]}{\exp (z_j \cdot \tilde{z}_a / \kappa)}}}  
     \vspace{-0.3cm}
\end{equation}
where $z_j = g_{\boldsymbol{w}_p}(e_j)$, $\tilde{z}_p = g_{\tilde{\boldsymbol{w}}_p}(\tilde{e_p})$ are projected student/teacher features, $\hat{P}(j) \triangleq \{p \in [\mathcal{Q}],  \max_m{(\tilde{q}_{p,m})} > \tau, \tilde{q}_{p} = q_j\}$ is the set of the indexes of weakly-augmented samples that have the same pseudo-label with $x_j$ and their confidences come up to $\tau$. 
We follow the designation of contrastive regularization \cite{lee2022contrastive}, while in our system the teacher model is instructed to guide the process. 
Specifically, the reference samples used for contrastive regularization are derived from the teacher features stored in the globally shared memory queue $\mathcal{Q}$, where features from prior supervised training are dequeued at a lower frequency.
Our total loss over unlabeled data is composed of the consistency regularization term and clustering regularization term: 
\vspace{-0.3cm}
\begin{equation}
    \ell_{u} = \mathcal{H} + \mathcal{C}
    \vspace{-0.3cm}
\end{equation}

By performing backward propagation, the estimated gradients for client $i$ computed as $\tilde{\nabla}_s f_{u, i}(\boldsymbol{w}_{s}^{h +, k}) = \frac{1}{|\mathcal{B}_{u,i}|}\sum_{x_{i,j} \in \mathcal{B}_{u,i}} {\nabla \ell_{ce}(x_{i,j}, \boldsymbol{w}_{s}^{h +,k})}$ and $\tilde{\nabla}_p f_{u, i}(\boldsymbol{w}_{p}^{h +,k}) = \frac{1}{|\mathcal{B}_{u,i}|}\sum_{x_{i,j} \in \mathcal{B}_{u,i}} {\nabla \ell_{c}(x_{i,j}, \boldsymbol{w}_{p}^{h +,k})}$ is kept until getting through all student features received before. 
Subsequently, the PS updates its server-side models with learning rate $\eta_h$ as:
\begin{equation}
\begin{split}
    & \boldsymbol{w}_{s}^{h +,k+1} = \boldsymbol{w}_{s}^{h +,k} - \eta_h \frac{1}{N} \sum\limits_{i \in [N]} \tilde{\nabla}_S f_{u, i}(\boldsymbol{w}_{s}^{h +, k}) \\
    & \boldsymbol{w}_{p}^{h +, k+1} = \boldsymbol{w}_{p}^{h +, k} - \eta_h \frac{1}{N} \sum\limits_{i \in [N]} \tilde{\nabla}_P f_{u, i}(\boldsymbol{w}_{p}^{h +,k})
\end{split}
\label{server_split_update_eq}
\vspace{-0.2cm}
\end{equation}

Along with that, the gradients of student features $\boldsymbol{e}_i$ computed as $d \boldsymbol{e}_i = \{ \nabla \ell_{u}(e_{i,j \in [\mathcal{B}_{u,i}]}, \boldsymbol{w}_{c, i}^{h +,k}) \}$ with a batch size of $d_i^{h +,k}$ are sent to the corresponding client $i$. 
Then each client continues to perform backward propagation $\tilde{\nabla}_c f_{u, i}(\boldsymbol{w}_{c, i}^{h +,k}) = \frac{1}{|\mathcal{B}_{u,i}|}\sum\limits_{j \in [\mathcal{B}_{u,i}]} {\nabla \ell_{u}(e_{i,j}, \boldsymbol{w}_{c, i}^{h +,k})}$ and separately updates its student/teacher bottom model as:
\begin{equation}
\begin{split}
    & \boldsymbol{w}_{c,i}^{h +,k+1} = \boldsymbol{w}_{c,i}^{h +,k} - \eta_h \tilde{\nabla}_c f_{u, i}(\boldsymbol{w}_{c, i}^{h +,k}) \\
    & \tilde{\boldsymbol{w}}_{c,i}^{h +,k+1} = \gamma \tilde{\boldsymbol{w}}_{c,i}^{h +,k} + (1-\gamma) \boldsymbol{w}_{c,i}^{h +,k+1}
\end{split}
\label{client_split_update_eq}
\end{equation}


(5) \textbf{\textit{Bottom Model Aggregation}.} 
After total $K_u$ iterations, clients upload their bottom models to the PS at a synchronization barrier, while the teacher bottom models, which are obtained from \srvUp, are excluded from model uploading. 
The server-side models are set as $\boldsymbol{w}_{s}^{h+1} = \boldsymbol{w}_{s}^{h +, K_u + 1}$ and $\boldsymbol{w}_{p}^{h+1} = \boldsymbol{w}_{p}^{h +, K_u + 1}$. 
The PS aggregates the bottom models uploaded by clients to obtain a global bottom model $\boldsymbol{w}_{c}^{h+1} = \frac{1}{N} \sum_{i \in [V]}{ \boldsymbol{w}_{c,i}^{h +, K_u + 1}}$, which, together with the modules residing at the PS, is assembled for further \srvUp.

\vspace{0.2cm}
\section{Algorithm Design}\label{algdesign}
In this section, we propose a greedy algorithm for \InvReg frequency adaptation, which is an important component of our \method system.
We first analyze the convergence bound after $H$ rounds \textit{w.r.t.} both the \InvReg frequency and the \Reg frequency. 
Then we would present our algorithm design and explain how it contributes to model convergence.

\begin{figure}[t]
    \centering
    \captionsetup{justification=centering}
    \subfloat[Semi-supervised training] {
    \begin{minipage}[t]{0.23\textwidth}
        \centering
        \includegraphics[scale=0.27]{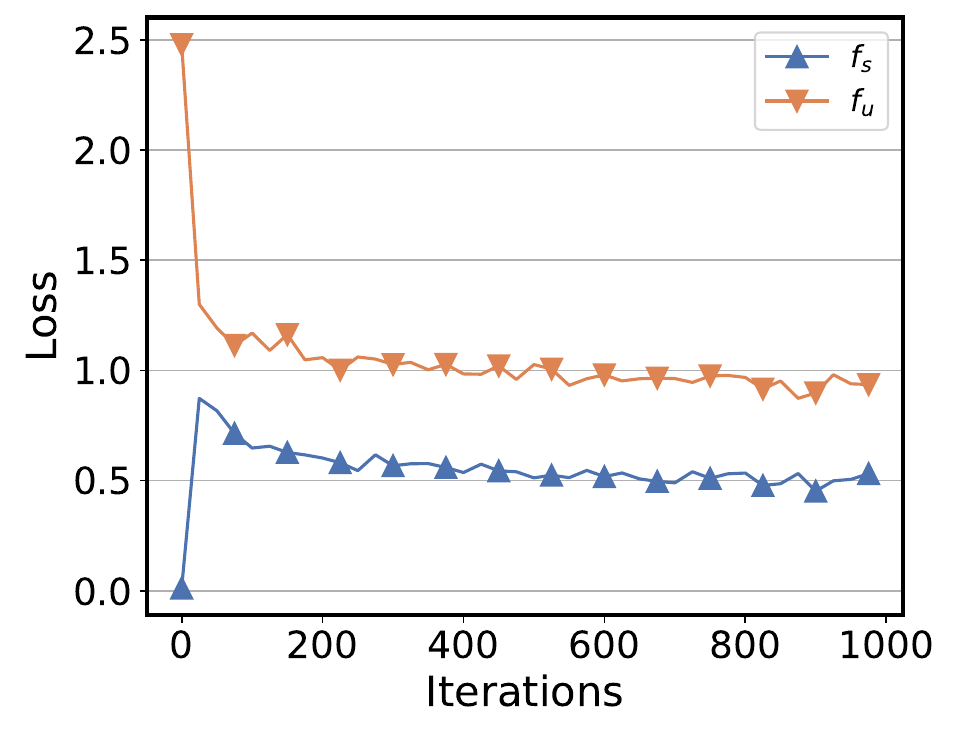}   
    \end{minipage}
    }
    \subfloat[Supervised training] {
    \begin{minipage}[t]{0.23\textwidth}
        \centering
        \includegraphics[scale=0.27]{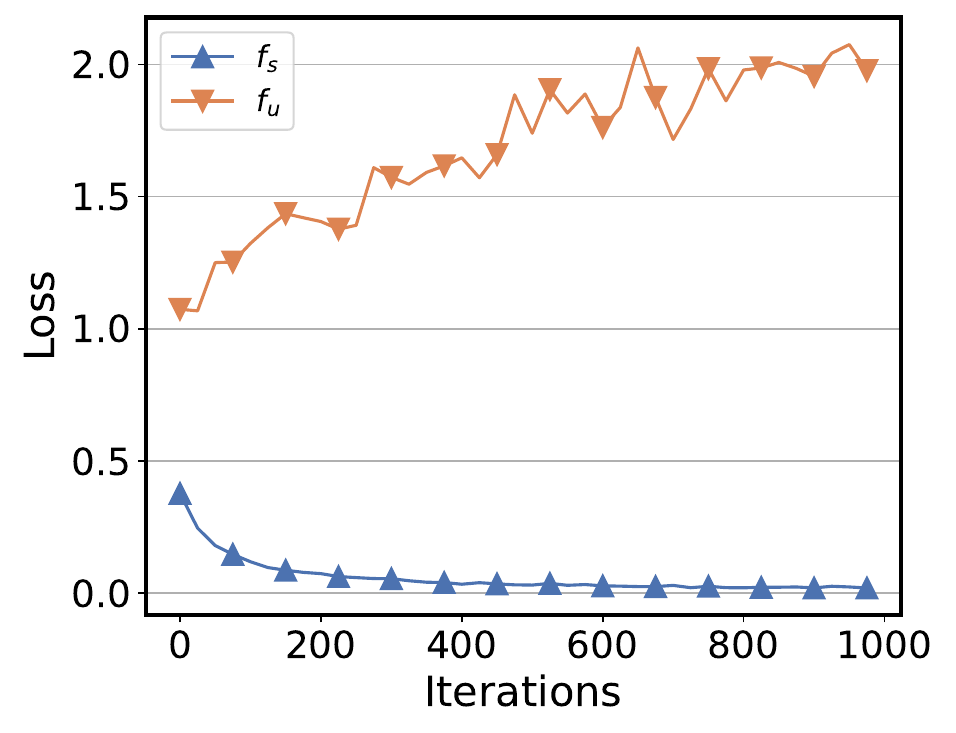}
    \end{minipage}
    }
    \vspace{-0.1cm}
    \caption{Loss variation on alternate training phases.}
    \vspace{-0.5cm}
    \label{fig-alg-lossv}
\end{figure}

\begin{figure}[t]
    \centering
    \captionsetup{justification=centering}
    \subfloat[Round \textit{vs.} $K_s$.] {
    \begin{minipage}[t]{0.23\textwidth}
        \centering
        \includegraphics[scale=0.27]{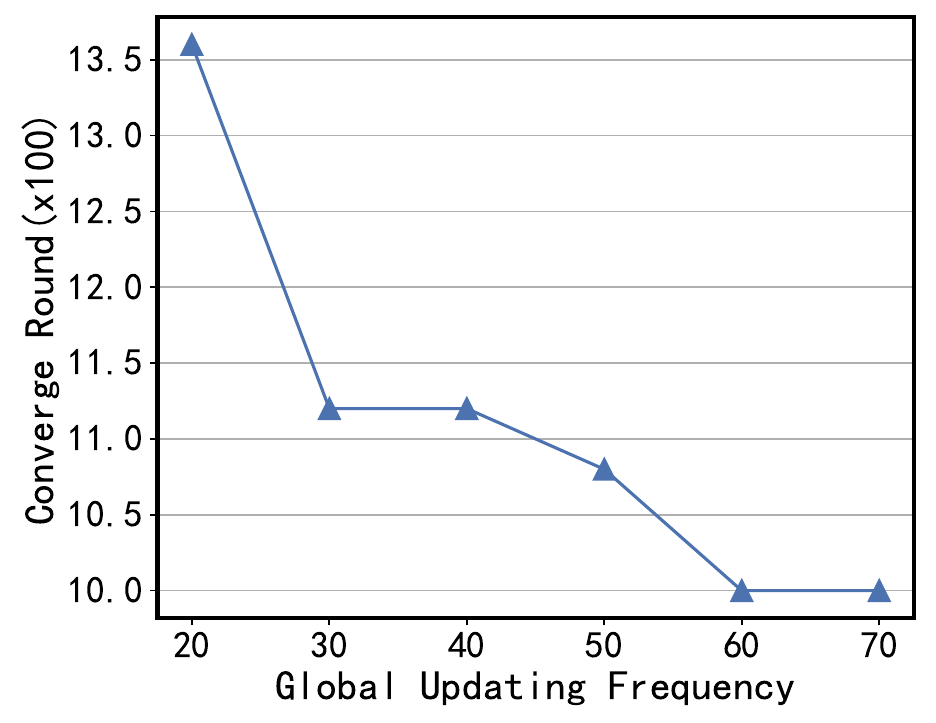}
    \end{minipage}
    }
    \subfloat[Accuracy \textit{vs.} $K_s$.] {
    \begin{minipage}[t]{0.23\textwidth}
        \centering
        \includegraphics[scale=0.27]{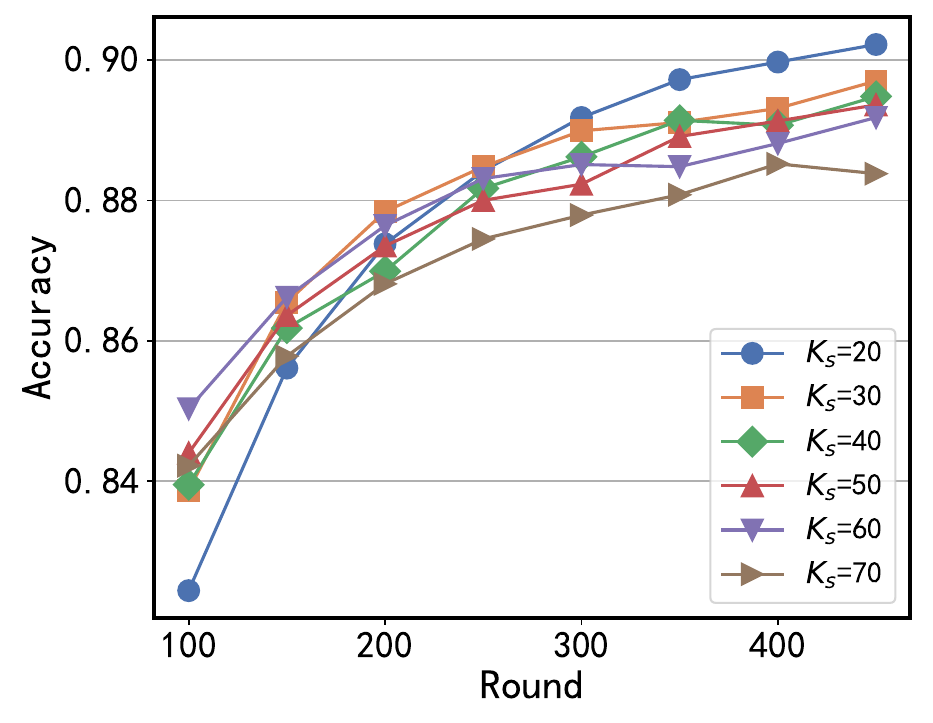}
    \end{minipage}
    }
    \vspace{-0.1cm}
    \caption{Model training with different $K_s$.}
    \vspace{-0.5cm}
    \label{fig-alg-sfreq}
\end{figure}

\subsection{Convergence Analysis}
\label{analysis-section}
Since the training is performed in an alternate manner, the convergence of the loss function relates to the training status in both the \srvUp and \locReg stages. 
For the sake of analysis, we make the following assumptions as suggested in \cite{li2019convergence, yang2021achieving, haddadpour2019convergence, ajalloeian2020convergence}:
\begin{assumption}
\label{assump1}
    (Lipschiz Continuous Gradient) The loss function $F(\cdot)$ and loss components $f_s(\cdot)$ and $f_{u,i}(\cdot)$ of the entire model are L-smooth such that:
    \begin{equation}
    \begin{split}
        & \Vert \nabla F(\boldsymbol{x}) - \nabla F(\boldsymbol{y}) \Vert \leq L \Vert \boldsymbol{x} - \boldsymbol{y} \Vert, \forall \boldsymbol{x}, \forall\boldsymbol{y};\\
        & \Vert \nabla f_s(\boldsymbol{x}) - \nabla f_s(\boldsymbol{y}) \Vert \leq L \Vert \boldsymbol{x} - \boldsymbol{y} \Vert, \forall \boldsymbol{x}, \forall\boldsymbol{y};\\
        & \Vert \nabla f_{u,i}(\boldsymbol{x}) - \nabla f_{u,i}(\boldsymbol{y}) \Vert \leq L \Vert \boldsymbol{x} - \boldsymbol{y} \Vert, \forall \boldsymbol{x}, \forall\boldsymbol{y}. \\
        \nonumber
    \end{split}
    \end{equation}
\end{assumption}

\begin{assumption}
\label{assump2}
\vspace{-0.6cm}
    \bluenote{(Bounded Second Moments \cite{han2021accelerating, yang2021achieving})} There exist constants $G_s$ and $G_u$, such that the second moments of the stochastic gradients of the unsupervised loss and supervised loss on any data sample are upper bounded by:
    \begin{equation}
    \begin{split}
        & \Vert \nabla \ell_{u}(x, \boldsymbol{w}) \Vert^2 \leq G_u^2, \forall x, \forall \boldsymbol{w}; \\
        \vspace{0.2cm}
        & \Vert \nabla \ell_{s}(x, \boldsymbol{w}) \Vert^2 \leq G_s^2, \forall x, \forall \boldsymbol{w}. \\  
        \nonumber
    \end{split}
    \end{equation}
\end{assumption}

\begin{assumption}
\label{assump3}
    \vspace{-0.6cm}
    (Unbiased Gradient Estimator) Let $\nabla f_s(\boldsymbol{w})$ and $\nabla f_{u,i}(\boldsymbol{w}_i)$ separately denote the gradients derived from the labeled data on the PS and from the unlabeled data of client $i$. The gradient estimators are unbiased as:
    \begin{equation}
    \begin{split}
        & \mathbb{E} \Vert \tilde{\nabla} f_{s}(\boldsymbol{w}) \Vert = \mathbb{E} \Vert \nabla f_{s}(\boldsymbol{w}) \Vert, \forall \boldsymbol{w};\\ 
        & \mathbb{E} \Vert \tilde{\nabla} f_{u, i}(\boldsymbol{w}) \Vert = \mathbb{E} \Vert \nabla f_{u, i}(\boldsymbol{w}) \Vert, \forall \boldsymbol{w}
        \nonumber.    
    \end{split}
    \end{equation}
\end{assumption}

Based on those assumptions, we have the following results given an initialized model $\boldsymbol{w}^0$.

\begin{theorem} 
\label{theorem1}
\vspace{-0.2cm}
The sequence of outputs $\{\boldsymbol{w}^{h, k}\}$ generated by supervised training and global aggregation satisfies:
\begin{equation}
 \underset{h \in [H], k \in \{0, \cdots, K_s\}}{\min} {\mathbb{E} \Vert \nabla F(\boldsymbol{w}^{h, k}) \Vert^2} 
    \leq \frac{2(F(\boldsymbol{w}^0) - F(\boldsymbol{w}^*))}{\eta_1 HK_s} 
    +\Phi.
    \nonumber
\end{equation}
where $\Phi \triangleq (\frac{L^2(K_u-1)(2K_u-1)\eta_1^2}{3K_s} + \frac{LK_u^2 \eta_1^3}{K_s} + 1) G_u^2 + (L \eta_1 + \frac{2 K_u}{K_s}) G_s^2$.
\bluenote{The detailed proof is presented in Appendix A.}
\end{theorem}

Theorem \ref{theorem1} suggests that the convergence of the model $\boldsymbol{w}$, obtained through alternating training, is intricately tied to $G_s^2$ and $G_u^2$. 
Specifically, as training goes on, continuous model updating either on \srvUp or \locReg stage ensures that the corresponding second-order moments, $G_s^2$ or $G_u^2$, trend towards zero. 
This indicates that model convergence is guaranteed if the updating frequency on either training stage nears zero. 
However, updating frequencies for both stages are often set the same in practice, resulting in a seesaw relationship between $G_s^2$ and $G_u^2$.
Thus, convergence is impeded due to fluctuations in $\Phi$.


\bluenote{
Next, we delve into a more detailed analysis based on
the experimental observations.
Theorem \ref{theorem1} highlights the complexity of jointly optimizing $K_u$ and $K_s$ to reduce the expected gradient of the loss function. 
Particularly, $K_u$ presents a multifaceted challenge as it impacts both the speed at which the model converges and the volume of features/gradients exchanged between clients and the PS per round, thereby affecting communication costs.
For the sake of simplification, we opt to keep $K_u$ constant and concentrate our efforts on exploring the effects of adjusting $K_s$.
We simulate a simple environment comprising 10 clients and 1 PS with $K_u=50$ and adjust $K_s$. 
We record the number of aggregation rounds needed to reach equilibrium  alongside the accuracy for each round. 
Equilibrium is defined by a variance in total loss falling below 0.01 across 50 consecutive rounds.
Echoing Theorem \ref{theorem1}, an increased global updating frequency facilitates a quicker transition of the model into a dynamic equilibrium, \ie, the expected gradient approaches to $G_u^2$ rapidly. 
This premise receives empirical support in Fig. \ref{fig-alg-sfreq}(a), where higher values of $K_s$ lead to convergence in fewer aggregation rounds.}

\bluenote{Additionally, within EC systems, where stringent time constraints prevail, the ideal global updating frequency might shift depending on available time budgets.
Fig. \ref{fig-alg-sfreq}(b) suggests that a higher $K_s$ accelerates the model's convergence to a higher accuracy state earlier on. 
On the flip side, a lower $K_s$ has the potential to yield greater accuracy during the latter training phases by nudging the equilibrium state closer to global optima (thereby reducing $\Phi$), albeit at the cost of more aggregation rounds, \ie, more resource consumption.
As such, we face a dilemma in determining the global updating frequency.}

\begin{figure}[t]
    \centering
    \captionsetup{justification=centering}
    \subfloat[$\Delta \Bar{f}_s$ \textit{vs.} $\Delta \bar{f}_u$ when $K_s=20$] {
    \begin{minipage}[t]{0.23\textwidth}
        \centering
        \includegraphics[scale=0.27]{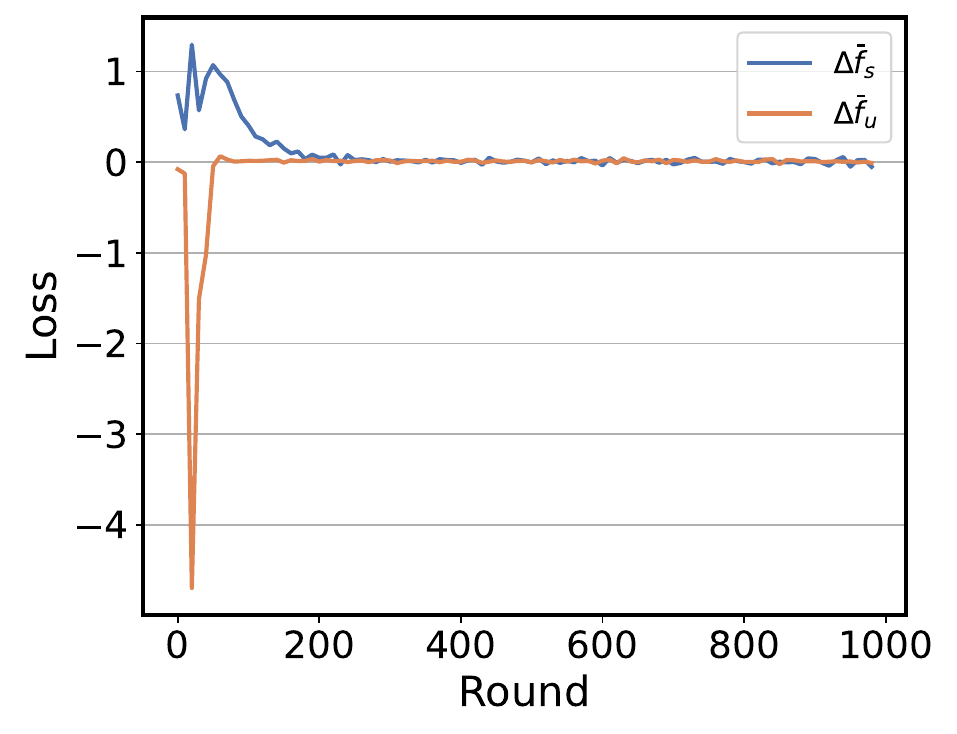}   
    \end{minipage}
    }
    \subfloat[$R_h$ with different $K_s$] {
    \begin{minipage}[t]{0.23\textwidth}
        \centering
        \includegraphics[scale=0.27]{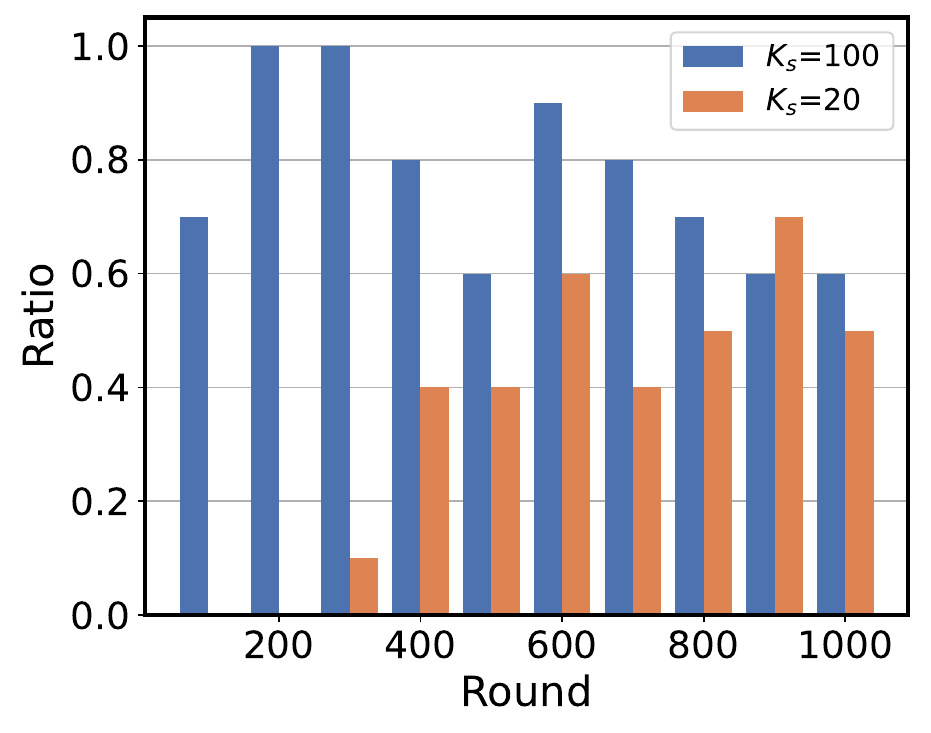}
    \end{minipage}
    }
    \vspace{-0.1cm}
    \caption{Loss Reduction during training.}
    \label{fig-alg-lossr}
    \vspace{-0.5cm}
\end{figure}

\begin{figure}[t]
    \centering
    \captionsetup{justification=centering}
    \subfloat[Loss Variation] {
    \begin{minipage}[t]{0.23\textwidth}
        \centering
        \includegraphics[scale=0.27]{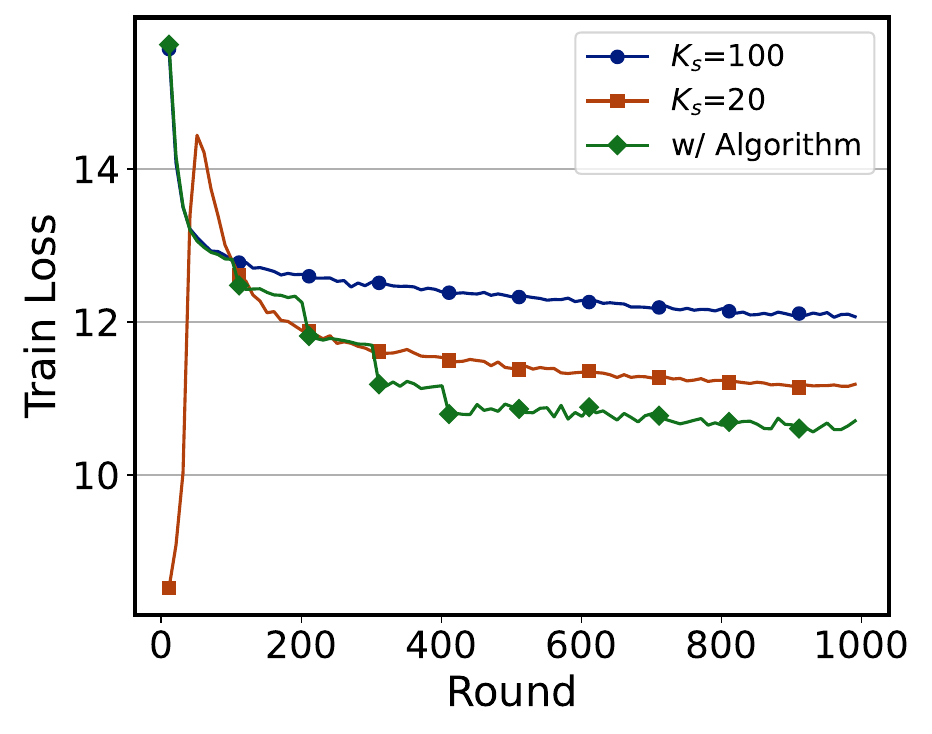}
    \end{minipage}
    }
    \subfloat[Accuracy Variation] {
    \begin{minipage}[t]{0.23\textwidth}
        \centering
        \includegraphics[scale=0.27]{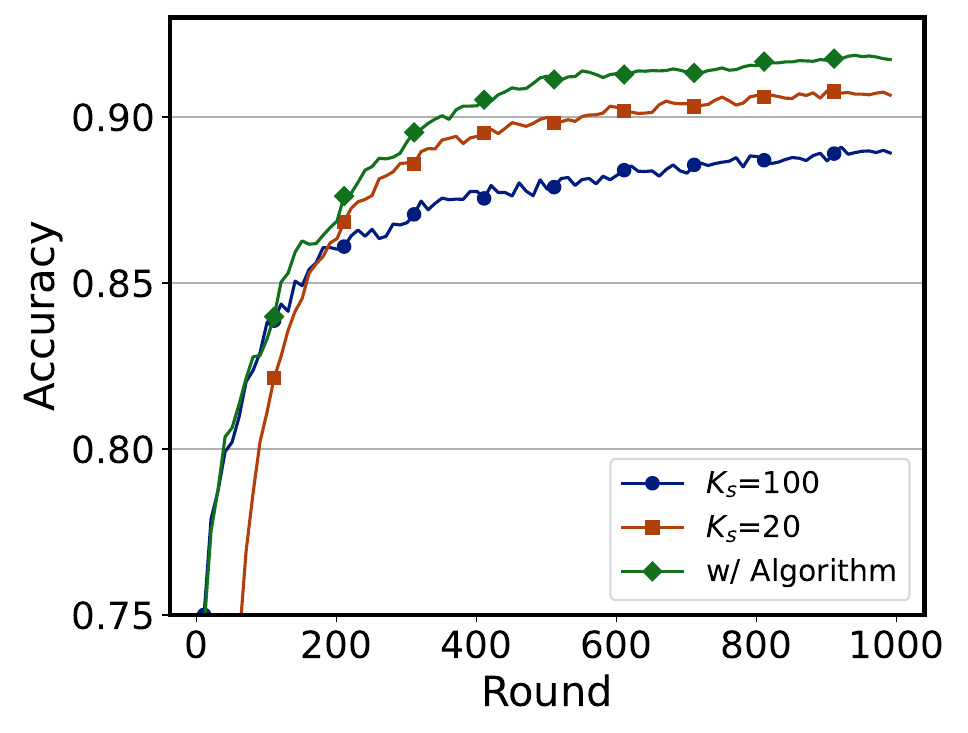}
    \end{minipage}
    }
    \vspace{-0.1cm}
    \caption{Model training with and without algorithm.}
    \label{fig-alg-train}
    \vspace{-0.6cm}
\end{figure}

\subsection{Global Updating Frequency Adaptation}
\vspace{-0.1cm}
In practical applications, it is impractical to determine an optimal global updating frequency in advance of training. 
To dynamically adjust the global updating frequency, we focus on analyzing the training dynamics. 
Generally, the model exhibits significant differences in training status between the initial and later training phases. 
Initially, the model parameters are initialized randomly and the supervised loss decreases rapidly, which denotes a large $\Delta f_s(\boldsymbol{w})$. 
Given the potential bias in pseudo labels, it is imperative to initialize a higher global updating frequency to prioritize the supervised training phase for keeping the model focused on correct targets. 
As training progresses, the supervised loss stabilizes, which indicates that the model has sufficiently assimilated the information from the labeled data. 
Then, the model should shift towards the semi-supervised training stage to minimize interference from semi-supervised training and expedite convergence towards the global optimal. 
Therefore, the crux lies in accurately seizing the moment to adjust the global updating frequency for balanced training.

\begin{algorithm}[t]
\caption{Training process on the PS}
\begin{algorithmic}[1]
    \STATE Initialize global model $\boldsymbol{w} = (\boldsymbol{w}_c, \boldsymbol{w}_s)$, teacher model $\tilde{\boldsymbol{w}} = \boldsymbol{w}$, $K_s^0 = K_s$
    \FOR{$h=0$ to $H-1$}
        \FOR{$k=0$ to $K_s^h-1$}
            \STATE Update $\boldsymbol{w}^{h, k+1} = \boldsymbol{w}^{h, k} - \eta_h\tilde{\nabla} f_s(\boldsymbol{w}^{h, k})$ 
            \STATE Update $\tilde{\boldsymbol{w}}^{h, k+1} = \gamma \tilde{\boldsymbol{w}}^{h, k} + (1-\gamma) \boldsymbol{w}^{h , k+1}$
        \ENDFOR
        \STATE Send $\boldsymbol{w}_c^{h+}$ and $\tilde{\boldsymbol{w}}_c^{h+}$ to all clients
        \STATE Set $\boldsymbol{w}_s^{h+,0} = \boldsymbol{w}_s^{h+}$
        \FOR{$k=1$ to $K_u$}
            \FOR{$i$ in $V$}
                \STATE Get ($\boldsymbol{z}_i$, $\tilde{\boldsymbol{z}}_i$) from client $i$
                \STATE Forward propagation with $\boldsymbol{z}_i$ and $\tilde{\boldsymbol{z}}_i$ on $\boldsymbol{w}_s^{h+,k}$ and  $\tilde{\boldsymbol{w}}_s^{h+,k}$
                \STATE Backward propagation, calculate $\tilde{\nabla}_s f_{u,i}(\boldsymbol{w}_s^{h+,k})$
                \STATE Send the gradients of $\boldsymbol{z}_i$ to client $i$
            \ENDFOR
            \STATE Update top models as Eq. (\ref{server_split_update_eq})
        \ENDFOR
        \FOR{$i$ in $V$}
            \STATE Get $\boldsymbol{w}_{c,i}^{h+, K_u}$ from client $i$
        \ENDFOR
        \STATE Set $\boldsymbol{w}_c^{h+1} = \frac{1}{N}\sum_{i \in [N]}\boldsymbol{w}_{c,i}^{h+, K_u}$
        \STATE Update $\Delta \bar{f}_s^h$ and $\Delta \bar{f}_u^h$ and calculate $R_h$
        \STATE Update $K_s^{h+1}$ as Eq. (\ref{eq-fdecay})
    \ENDFOR
\end{algorithmic}
\label{alg-process}
\end{algorithm}

To delve deeper into this matter, we conduct additional experiments on the CNN model described earlier.
In each round, we monitor both the supervised loss, $f_s(\boldsymbol{w}^h)$, and the semi-supervised loss, $f_u(\boldsymbol{w}^h)$, hereafter referred to simply as $f_s^h$ and $f_u^h$.
Following this, we calculate the mean loss across each discrete observational period---comprising 10 rounds---and label these as $\bar{f}_s^n$ and $\bar{f}_u^n$ (with $n$ extending from 1 to $\lceil \frac{H}{10} \rceil$).
Subsequently, we compute the change in average loss between successive observation periods, denoted as $\Delta \bar{f}_s^n$ and $\Delta \bar{f}_u^n$, which will be further examined over spans of 100 rounds.
For simplification, we define an indicator function as:
\vspace{-0.2cm}
\begin{equation}
  I_n = 
  \begin{cases} 
   1 & \text{if } \Delta \bar{f}_u^n > \Delta \bar{f}_s^n; \\
   0 & \text{otherwise}.
  \end{cases}
  \vspace{-0.2cm}
\end{equation}

As depicted in Fig. \ref{fig-alg-lossr}, during the initial training phase, the model is dominated by supervised loss ($I_n = 0$). 
As training progresses and output confidence grows, semi-supervised loss becomes more pronounced, evidenced by the proportion in every 100 rounds where $\Delta \bar{f}_u^n$ surpasses $\Delta \bar{f}_s^n$, represented by $R_h = \frac{1}{10}\sum_{n=1}^{10} I_n$. 
Notably, Fig. \ref{fig-alg-lossr} (b) unveils a significant phenomenon that when $K_s=100$, the initial $R_h$ is substantially higher compared to the scenario when $K_s=20$.
However, as training proceeds, this gap diminishes and may even invert.
This implies that semi-supervised training is negatively impacted sooner with a higher global updating frequency.
This insight, aligning with the pattern presented in Fig. \ref{fig-alg-train} (b), establishes $R_h$ as a crucial metric for evaluating the training progress.
Based on these observations, we propose a greedy algorithm for adaptively adjusting the global updating frequency with the following rule:
\vspace{-0.1cm}
\begin{equation}
\label{eq-fdecay}
K_{s}^{h+1}=
  \begin{cases}
    \max(\lfloor \frac{K_{s}^{h}}{\alpha}\rfloor, K_{min}) & \text{if } R_h \geq 0.5; \\
    K_{s}^{h} & \text{otherwise}.
  \end{cases}
  \vspace{-0.1cm}
\end{equation}
where $\alpha \in R^+$ is a decaying factor with $\alpha \geq 1$.

Next, we will explain our algorithm design.
Specifically, we start the training with a high global updating frequency ($K_s=100$) to encourage swift convergence. 
When the semi-supervised loss declines faster than the supervised loss, we adjust the global updating frequency downwards.
This allows the model to focus more on semi-supervised training in the later training phases by minimizing interference from supervised training. 
To mitigate the influence of random variations, adjustments to the global updating frequency are made only when more than half of the last 10 observation periods exhibit $\Delta \bar{f}_u^n > \Delta \bar{f}_s^n$.
Additionally, to prevent the model from overfitting to unlabeled data, we set a lower bound for the global updating frequency $K_{min}$ proportional to the ratio of labeled data available at the PS to the entire dataset, \ie, $K_{min} = \lfloor \beta\frac{|\mathcal{D}l|} {|\mathcal{D}|} K_u \rfloor$.
Finally, confirmatory experiments are conducted under the aforementioned system to validate the efficacy of our algorithm. 
Upon detecting $R_h \geq 0.5$, the algorithm proportionately reduces the global update frequency ($\alpha=2$), until it reaches the predefined minimum $K_{min}$($\beta=4$). 
These adjustments lead to a consistent reduction in overall loss, as depicted in Fig. \ref{fig-alg-train} (a), ensuring that our algorithm achieves the best possible model performance across varying computational constraints (Fig. \ref{fig-alg-train} (b)). 
In essence, our algorithm maintains a favorable balance between alternate training stages, thereby steering the model towards peak performance.

We present the overall training process on the PS, including the adaptation algorithm, in Alg. \ref{alg-process}. 
The training starts with supervised training, and the teacher model is updated batch-wise at the PS in each round for $K_s^h$ iterations (Lines 4-5 in Alg. \ref{alg-process}). 
The PS distributes the global bottom model and teacher bottom model to each client (Line 7 in Alg. \ref{alg-process}). 
The \locReg on PS is performed for each client in turn using the top model.
Then, the calculated gradients for student features (features generated by student bottom models) are sent to the corresponding clients (Lines 11-14 in Alg. \ref{alg-process}). 
After $K_u$ steps of \locReg, the bottom models are aggregated (Lines 18-21 in Alg. \ref{alg-process}). 
Finally, the PS updates $K_s$ based on the loss variation at the end of each round (Lines 22-23 in Alg. \ref{alg-process}) and starts the next round. 
The detailed of client training process have been described in Section \ref{system-design}, which includes conducting local forward propagation, uploading features, and performing backward propagation based on the downloaded gradients.

\section{Experiments and Evaluation}\label{evaluation}
In this section, we begin by providing a list of the datasets and models, as well as a description of the devices utilized in the experiments. 
We then introduce the adopted baselines and metrics for performance comparison. 
Finally, we present our evaluation results and analyze the superiority of \method under various scenarios.

\vspace{-0.1cm}
\subsection{Datasets and Models}
\textbf{Datasets:} We conduct experiments on four commonly used real-world datasets for semi-supervised learning: SVHN \cite{netzer2011reading}, CIFAR-10 \cite{krizhevsky2009learning}, STL-10 \cite{coates2011analysis} and IMAGE-100. 
The SVHN dataset contains 73,257 digits for training and 26,032 digits for testing, which are labeled in 10 classes. 
As suggested in \cite{diao2022semifl}, 1,000 labeled digits in the training set are allocated for the PS, while the remaining training data are distributed to clients as unlabeled data.
The CIFAR-10 dataset is an image dataset consisting of 60,000 32$\times$32 color images (50,000 for training and 10,000 for testing) in 10 categories, and 4,000 images are set as labeled data on the PS. 
The STL-10 dataset contains images from 10 different classes.
It includes 5,000 labeled images for training and 8,000 images for testing, with each image of size 96$\times$96. An additional 100,000 unlabeled images are provided for unsupervised learning, making this dataset particularly suitable for semi-supervised learning scenarios.
To evaluate \method on a more challenging task, we create the IMAGE-100 dataset, which is a subset of ImageNet \cite{russakovsky2015imagenet} and contains 100 out of 1,000 categories. 
Each sample in IMAGE-100 is resized to the shape of 144$\times$144$\times$3, and 5,000 labeled images are allocated for the PS.

\textbf{Models:} Four models with different types and structures are adopted on the above three real-world datasets for performance evaluation: (i) CNN on SVHN, (ii) AlexNet \cite{krizhevsky2017imagenet} on CIFAR-10, (iii) VGG13 \cite{simonyan2014very} on STL-10, (iv) VGG16 \cite{simonyan2014very} on IMAGE-100. 
For SVHN, we train a customized CNN model, which has two 5$\times$5 convolutional layers, a fully-connected layer with 512 units, and a softmax output layer with 10 units. 
For CIFAR-10, we train the AlexNet model with size of 127MB, composed of three 3$\times$3 convolutional layers, one 7$\times$7 convolutional layer, one 11$\times$11 convolutional layer, two fully-connected hidden layers, and one softmax output layer. 
For STL-10, the VGG-13 model with size of 508MB is adopted, which features 10 convolutional layers with 3$\times$3 kernel sizes, two fully-connected layers, and a softmax output layer for classifying images into 10 categories.
Lastly, the VGG16 model with 0.13B parameters and size of 528MB, which consists of 13 convolutional layers with kernel size of 3$\times$3, two fully-connected layers and a softmax output layer, is trained for image classification of IMAGE-100.

\begin{figure*}[t]
\centering
\captionsetup{justification=centering}
\subfloat[SVHN]{
\label{fig-timecost-SVHN}
\begin{minipage}[t]{0.24\linewidth}
    \centering
    \includegraphics[scale=0.3]{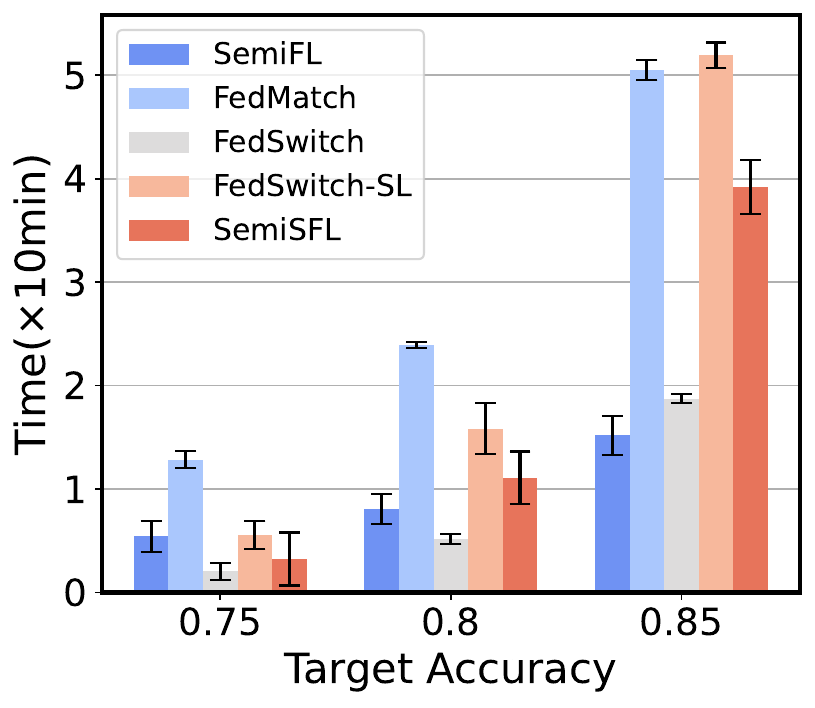}
\end{minipage}
}
\subfloat[CIFAR-10]{
\begin{minipage}[t]{0.24\linewidth}
    \centering
    \includegraphics[scale=0.3]{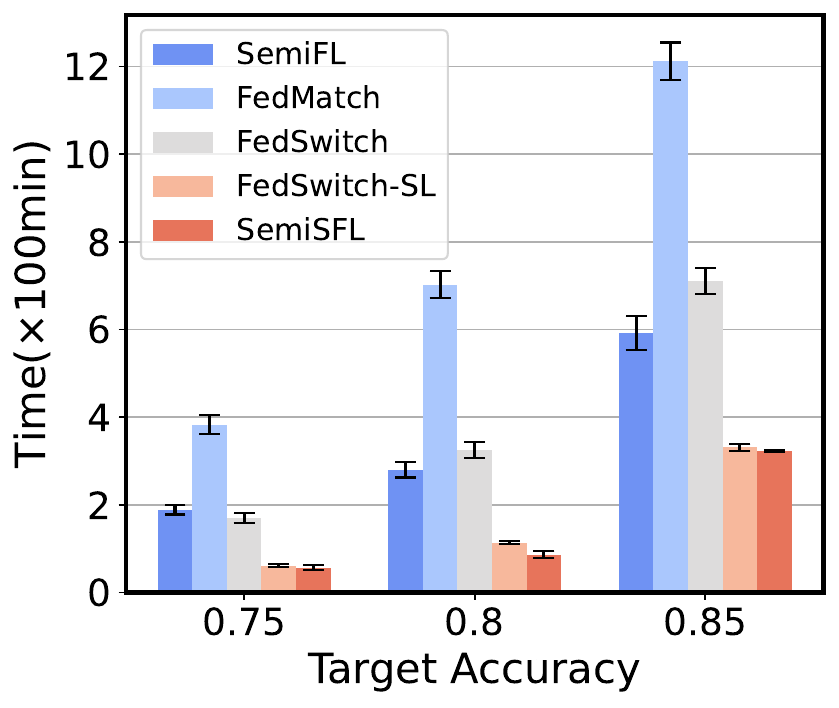}
\end{minipage}
}
\subfloat[STL-10]{
\begin{minipage}[t]{0.24\linewidth}
    \centering
    \includegraphics[scale=0.3]{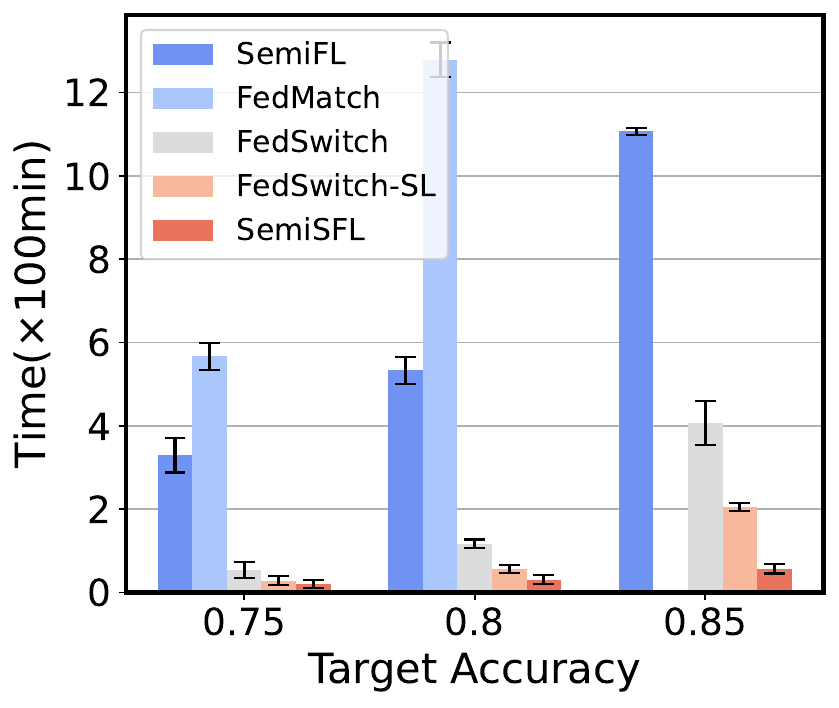}
\end{minipage}
}
\subfloat[IMAGE-100]{
\begin{minipage}[t]{0.24\linewidth}
    \centering
    \includegraphics[scale=0.3]{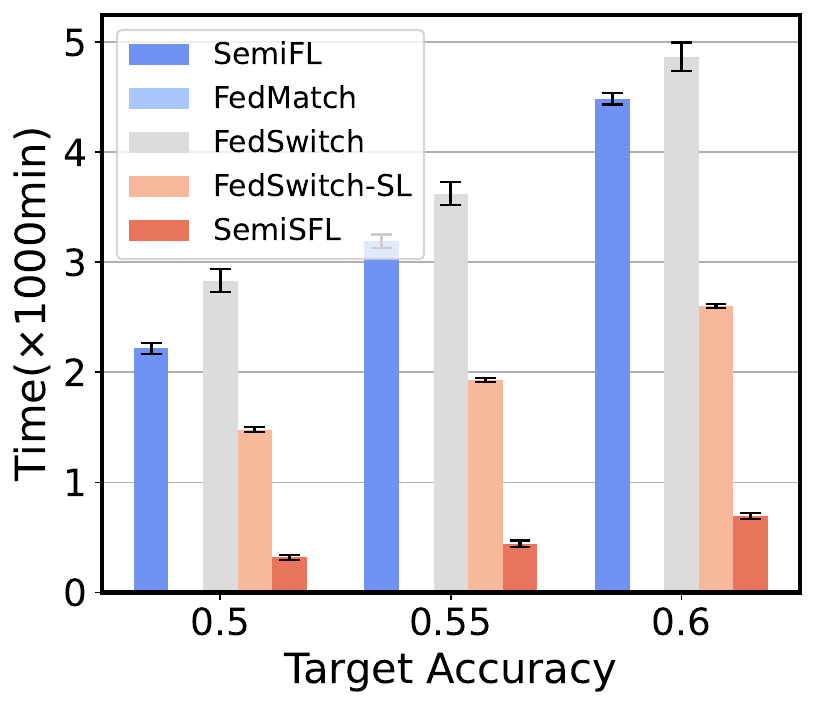}
\end{minipage}
}
\caption{Comparison of time cost on four datasets..\label{fig-timecost}}
\vspace{-0.6cm}
\end{figure*}

\subsection{Baselines and Metrics}
\textbf{Baselines:}  We measure the effectiveness of \method through a comparison with five baselines.
\begin{itemize}
\item[\textbullet] \textbf{Supervised-only}: 
Supervised-only refers to using only the labeled dataset available on the PS for supervised training. 
This represents the lower bound that can be achieved with a limited amount of labeled data.


\item[\textbullet] \textbf{SemiFL} \cite{diao2022semifl}: 
SemiFL is the first approach for semi-supervised FL with accuracy on par with standalone training, in which pseudo-labels for the local data of each client are generated upon the latest global model.
In a certain iteration, each client applies the Mixup \cite{berthelot2019mixmatch} technique on each data batch to augment the data and then perform forward and backward propagation using a specialized ``mix'' loss.

\item[\textbullet] 
\textbf{FedMatch} \cite{jeong2021federated}: 
FedMatch learns inter-client consistency by mutual sharing of client models. 
To mitigate interference between supervised and semi-supervised tasks, FedMatch decomposes model parameters into two variables, which are updated in an alternate way.

\item[\textbullet] 
\textbf{FedSwitch} \cite{zhao2023does}: 
FedSwitch is the state-of-the-art approach that leverages an EMA model, \ie, a teacher model, to ensure the quality of pseudo-labels. 
Additionally, it adaptively switches between the teacher and student model for pseudo-labeling, both to enhance the quality of pseudo-labels in non-IID settings and reduce communication costs.

\item[\textbullet]
\textbf{FedSwitch-SL}: An extension version of FedSwitch that incorporates the technique of Split Learning (SL). It serves as an ablation study to demonstrate the effectiveness of our clustering regularization.

\end{itemize}

\textbf{Metrics:} We employ the following metrics to evaluate the performance of different approaches.
\begin{itemize}
\item[\textbullet] \textbf{Test accuracy}: In each round, we measure the accuracy of the global model on the test set with different baselines. 
For FedSwitch(-SL) and \method, we use the global teacher model for testing.
\item[\textbullet] \textbf{Time cost}: We record the total time taken to achieve target test accuracy on different approaches, which includes the time for computation and communication.
\item[\textbullet] \textbf{Communication cost}: The communication cost for transmitting models and feature batches (if any) between entities to achieve the target accuracy is also recorded.
\end{itemize}

\subsection{Experimental Setup}
We evaluate the performance of \method on a hardware prototype system, which consists of an AMAX deep learning workstation equipped with an Intel(R) Xeon(R) Platinum 8358P CPU, 8 NVIDIA GeForce RTX A6000 GPUs, and 512 GB RAM, along with 80 Nvidia Jetson devices.
To represent the heterogeneity among clients, the NVIDIA Jetson devices operate under different modes, with various configurations of active CPUs and varying CPU/GPU frequencies \cite{liao2023accelerating, xu2024overcoming}.
Note that the algorithms and theoretical analyses presented in this paper are applicable to a variety of edge devices with diverse computing and communication capabilities, beyond the specific devices mentioned.




Additionally, the clients and the PS are connected via wireless links. 
Due to random channel noise and varying distances between devices and the router, the outbound bandwidth fluctuates between 0.8 Mbps and 8 Mbps, and the inbound bandwidth varies between 10 Mbps and 20 Mbps.
We use the PyTorch deep learning framework for our software implementation of model training and build up the connections between clients and the PS using the socket library.
The source code is available at \url{https://github.com/littlefishe/Capsule}.

We run 10 trials for all benchmark models and datasets with different random seeds. 
The standard errors are presented in the tables and the figures.
By default, each experiment is run for 1,000 aggregation rounds on SVHN, CIFAR-10, STL-10, and IMAGE-100 to ensure convergence. 
Similar to \cite{diao2022semifl}, we use an SGD optimizer to optimize the models and adopt the cosine learning rate decay schedule \cite{loshchilov2016sgdr}. 
Besides, we set the same hyperparameters as in \cite{diao2022semifl}, where the optimizer momentum $\beta_l = 0.9$, the initial learning rate $\eta = 0.02$, and the confidence threshold $\tau = 0.95$.
As for the adaptation algorithm, we set $\alpha = 1.5$ and $\beta = 8$. 
For FedSwitch-SL and \method, the indices of the split layer we select are 2, 5, 10, and 13 for CNN, AlexNet, VGG13, and VGG16, respectively. 
Unless otherwise specified, the number of labeled data samples on the PS is 1,000, 4,000, 5,000, and 5,000 for SVHN, CIFAR-10, STL-10, and IMAGE-100, respectively, and the unlabeled data samples are distributed uniformly across clients.



\begin{table}[t]
\caption{Overall test accuracy (\%).}
\label{tab-ovacc}
\renewcommand{\arraystretch}{1.3} 
\begin{tabular}{lcccc}
\toprule
\multicolumn{1}{c}{\multirow{2}{*}{Baselines}} & \multicolumn{4}{c}{Dataset}                \\ \cline{2-5} 
\multicolumn{1}{c}{}                           & SVHN      & CIFAR-10  & STL-10 & IMAGE-100 \\
\midrule
Supervised-only                                & 73.6(0.2) & 75.1(0.5) & 81.9(0.2) & 26.1(0.7) \\
SemiFL                                         & 88.1(0.7) & 86.8(1.2) & 90.7(2.5) & 65.4(2.0) \\
FedMatch                                       & 88.4(0.6) & 86.1(1.1) & 77.5(1.1) & 29.6(1.7) \\
FedSwitch                                      & 89.0(0.5) & 87.9(1.0) & 91.5(0.3)    & 60.3(1.5) \\
FedSwitch-SL                                   & 89.1(0.4) & 87.5(0.8) & 90.1(0.4)    & 61.1(1.4) \\
\method                                          & \textbf{91.4(0.4)} & \textbf{88.6(0.2)} & \textbf{92.8(0.3)} & \textbf{66.6(0.3)} \\
\bottomrule
\end{tabular}
\vspace{-0.4cm}
\end{table}

\subsection{Experiment Results}

\begin{figure*}[t]
\centering
\captionsetup{justification=centering}
\subfloat[SVHN]{
\begin{minipage}[t]{0.24\linewidth}
    \centering
    \includegraphics[scale=0.3]{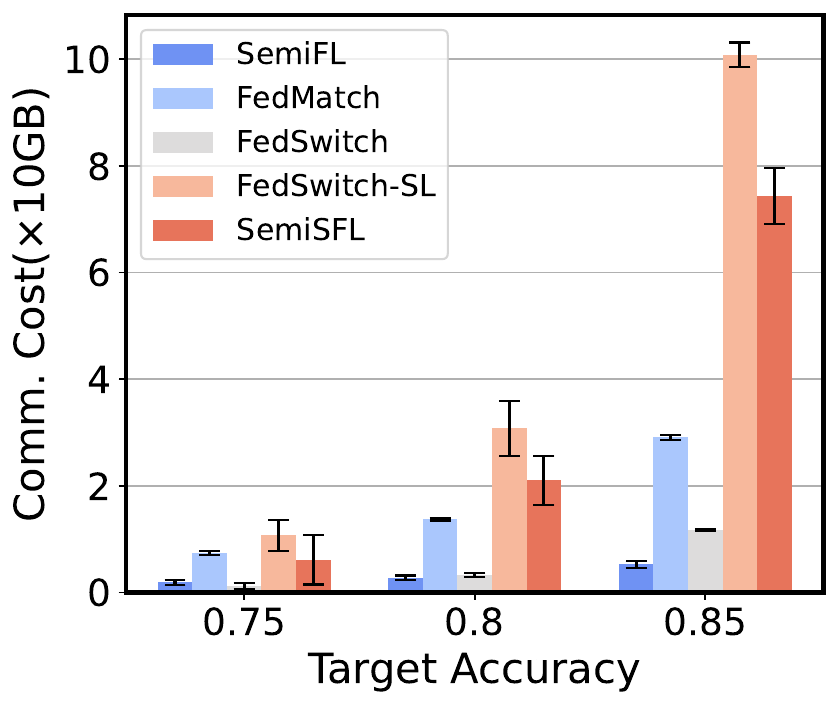}
\end{minipage}
}
\subfloat[CIFAR-10]{
\begin{minipage}[t]{0.24\linewidth}
    \centering
    \includegraphics[scale=0.3]{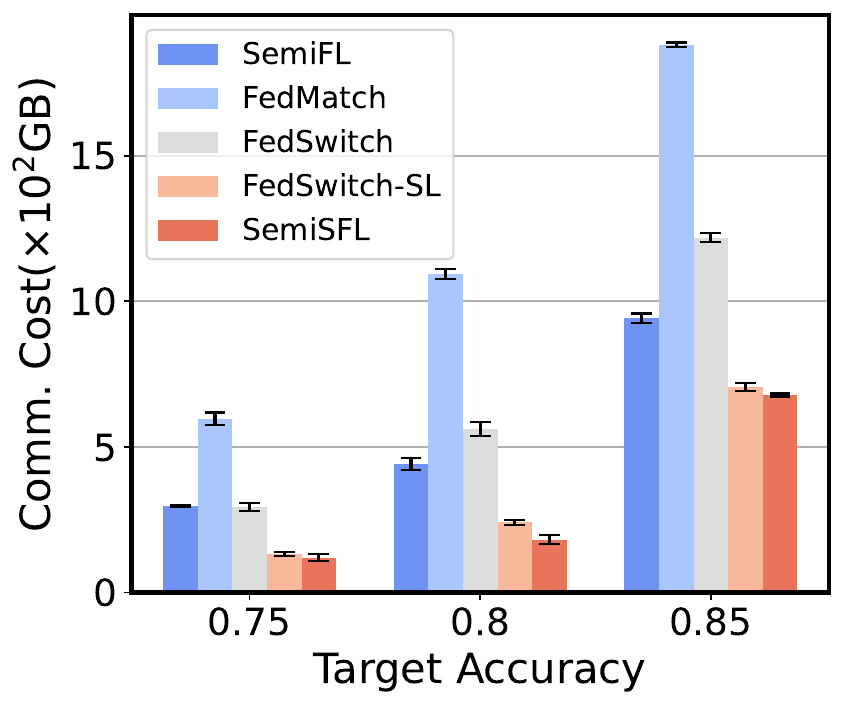}
\end{minipage}
}
\subfloat[STL-10]{
\begin{minipage}[t]{0.24\linewidth}
    \centering
    \includegraphics[scale=0.3]{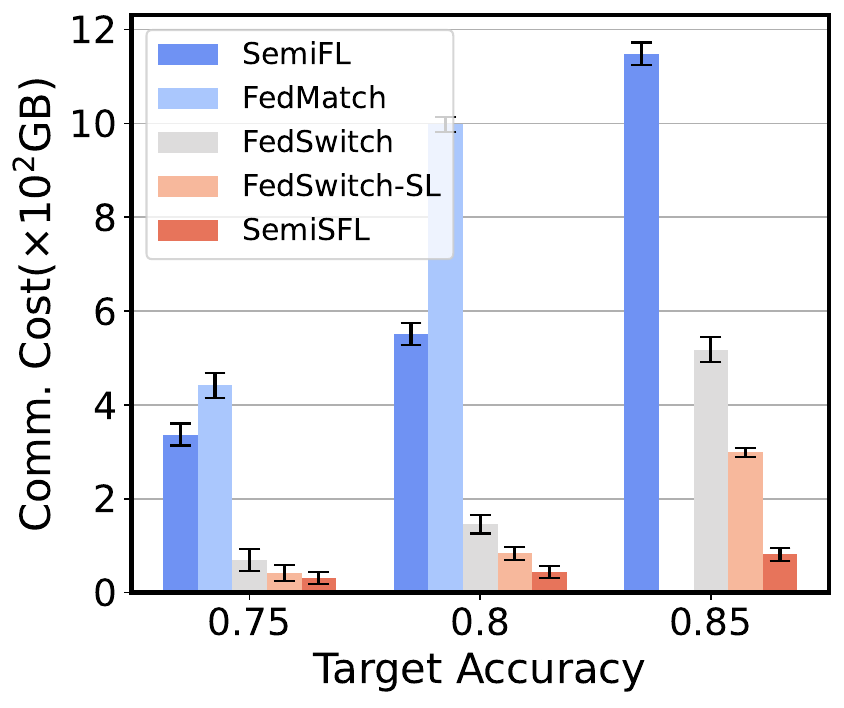}
\end{minipage}
}
\subfloat[IMAGE-100]{
\begin{minipage}[t]{0.24\linewidth}
    \centering
    \includegraphics[scale=0.3]{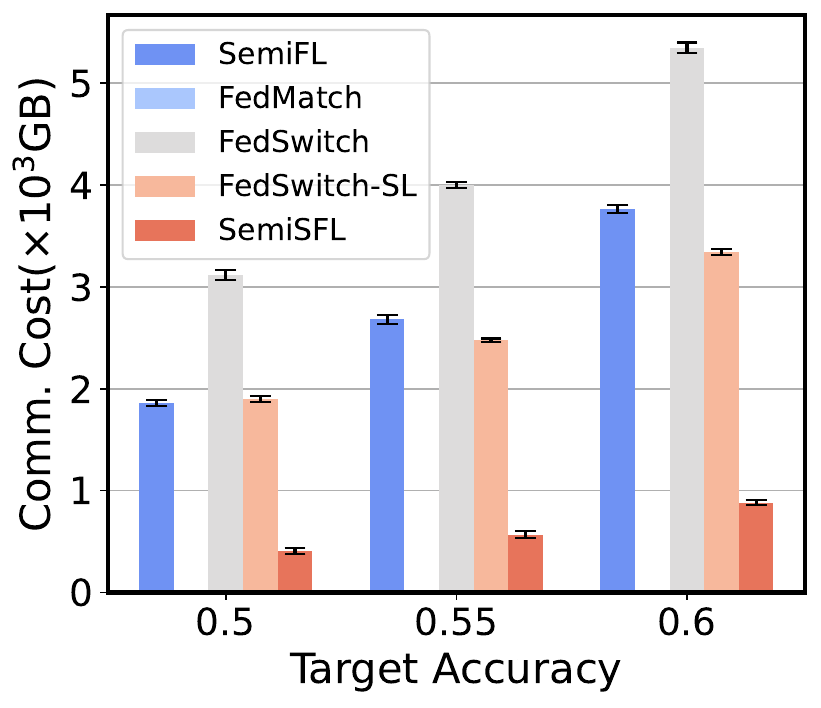}
\end{minipage}
}
\vspace{-0.15cm}
\caption{Comparison of communication cost on four datasets.\label{fig-commcost}}
\vspace{-0.4cm}
\end{figure*}

\begin{figure*}[t]
\centering
\captionsetup{justification=centering}
\subfloat[Dir(1.0)] {
\begin{minipage}[t]{0.24\linewidth}
    \centering
    \includegraphics[scale=0.3]{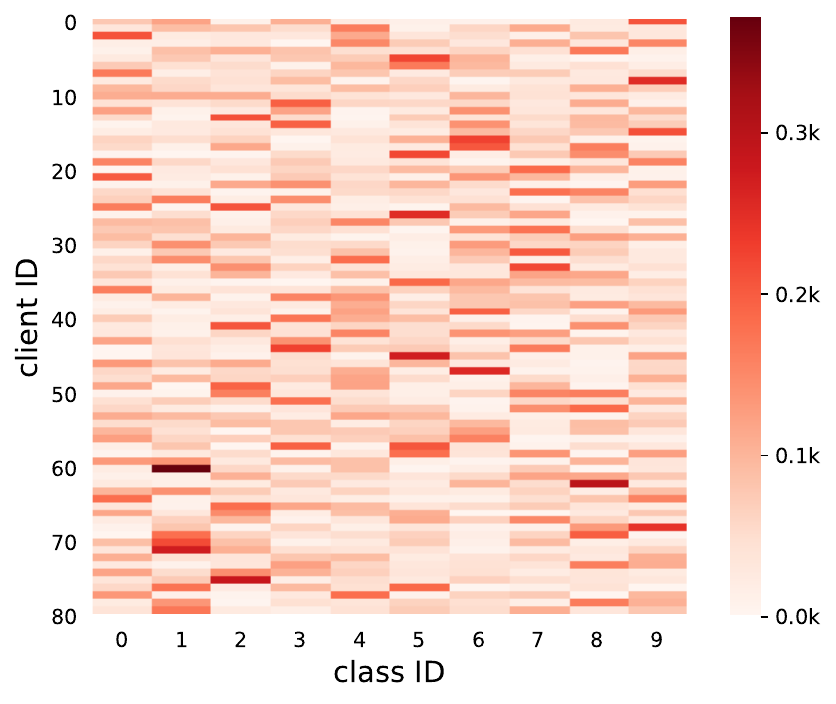}
\end{minipage}
}
\subfloat[Dir(0.5)] {
\begin{minipage}[t]{0.24\linewidth}
    \centering
    \includegraphics[scale=0.3]{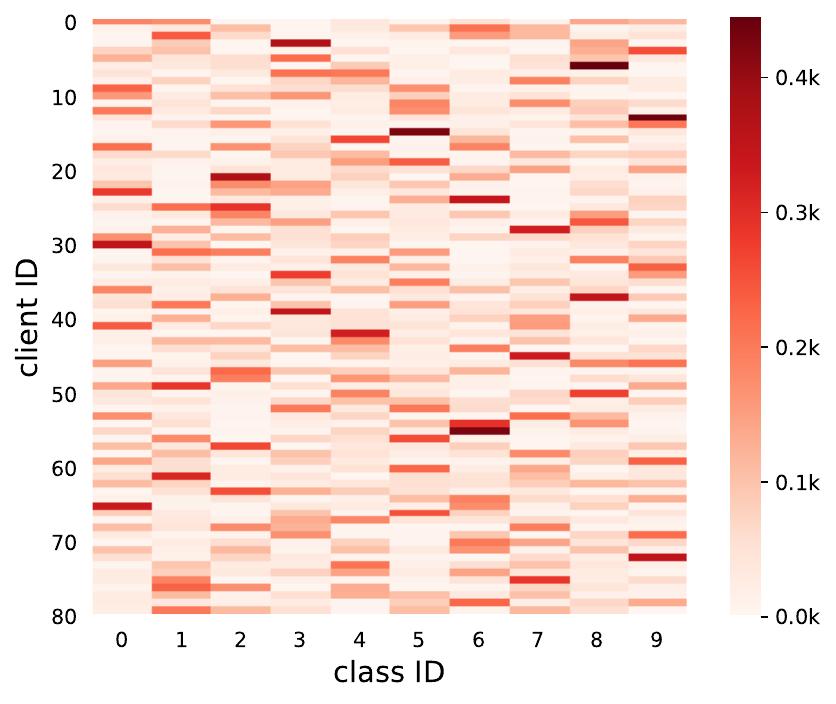}
\end{minipage}
}
\subfloat[Dir(0.1)] {
\begin{minipage}[t]{0.24\linewidth}
    \centering
    \includegraphics[scale=0.3]{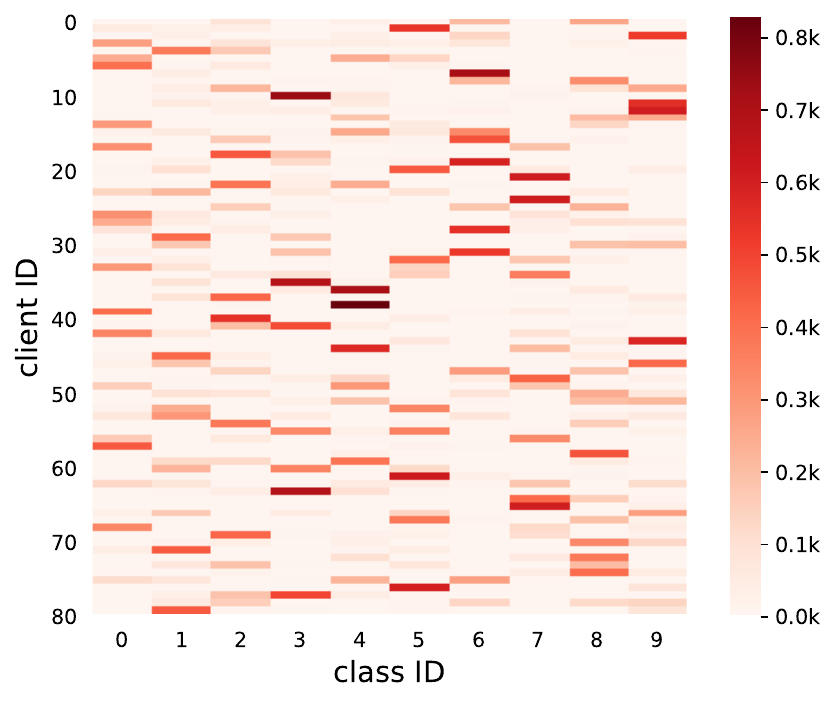}
\end{minipage}
}
\subfloat[Dir(0.05)] {
\begin{minipage}[t]{0.24\linewidth}
    \centering
    \includegraphics[scale=0.3]{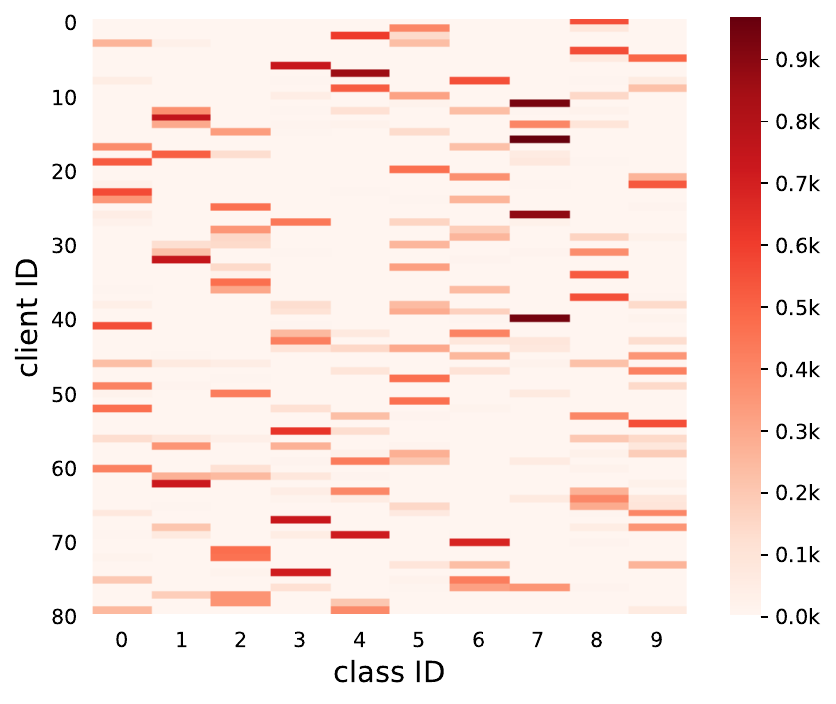}
\end{minipage}
}
\vspace{-0.1cm}
\caption{Data distribution of each client with different non-IID levels.}
\label{fig-hm}
\vspace{-0.5cm}
\end{figure*}



\subsubsection{Overall Effectiveness} 
The final test accuracies on SVHN, CIFAR-10, STL-10, and IMAGE-100 for \method and baselines are presented in Table \ref{tab-ovacc}. 
Supervised-only is omitted in the later sections for the comparison of training efficiency since the clients are not involved in this approach.
In terms of accuracy, \method consistently achieves the highest accuracy compared to state-of-the-art approaches. 
For instance, on the SVHN dataset, \method can achieve 91.4\% accuracy, which is 2.3\% higher than that in FedSwitch-SL. 
Similarly, \method achieves 88.6\% and 92.8\% accuracy on CIFAR-10 and STL-10, respectively, surpassing SemiFL by 1.8\% and 2.7\%.
Moreover, on IMAGE-100, \method can achieve 66.6\% accuracy, which exhibits a 5.5\% improvement over FedSwitch-SL.
\method also improves accuracy from 2.5\% to 37.0\% compared to FedMatch across the four datasets, highlighting the effectiveness of the clustering regularization.


We also present the time cost required to achieve different test accuracies in Fig. \ref{fig-timecost}. 
FedMatch is excluded from the comparison plots on IMAGE-100 and from some bars in the STL-10 plots because it fails to achieve the target accuracy.
When training small-scale models, \method fails to outperform all of the baselines on the SVHN dataset in terms of the time cost, as shown in Fig. \ref{fig-timecost}(a).
This happens when the cost of transmitting the features outweighs that of the entire model, as evidenced by the high overhead of the FedSwitch-SL.
However, both FedSwitch-SL and \method benefit from scaling the size of the model. 
For instance, to achieve 80\% accuracy on CIFAR-10, \method requires 86.5 min on average to train an AlexNet, compared to 113.6 min for FedSwitch-SL and 279.9-702.4 min for other FL baselines, indicating speed-ups of 1.3-8.1$\times$ over the baselines.
For a higher target accuracy of 85\%, \method only takes 322.8 min, while FedSwitch-SL requires 330.9 min, and other Semi-FL baselines span from 591.9 to 1211.9 min. 
Additionally, \method speeds up training by 3.6-19.5$\times$ to reach 85\% accuracy with VGG13 on STL-10 and by 3.8-7.0$\times$ to reach 60\% accuracy with VGG16 on IMAGE-100 compared to FedSwitch-SL and other Semi-FL baselines.

\begin{figure*}[t]
\centering
\captionsetup{justification=centering}
\subfloat[Dir(1.0)]{
\begin{minipage}[t]{0.24\linewidth}
    \centering
    \includegraphics[scale=0.3]{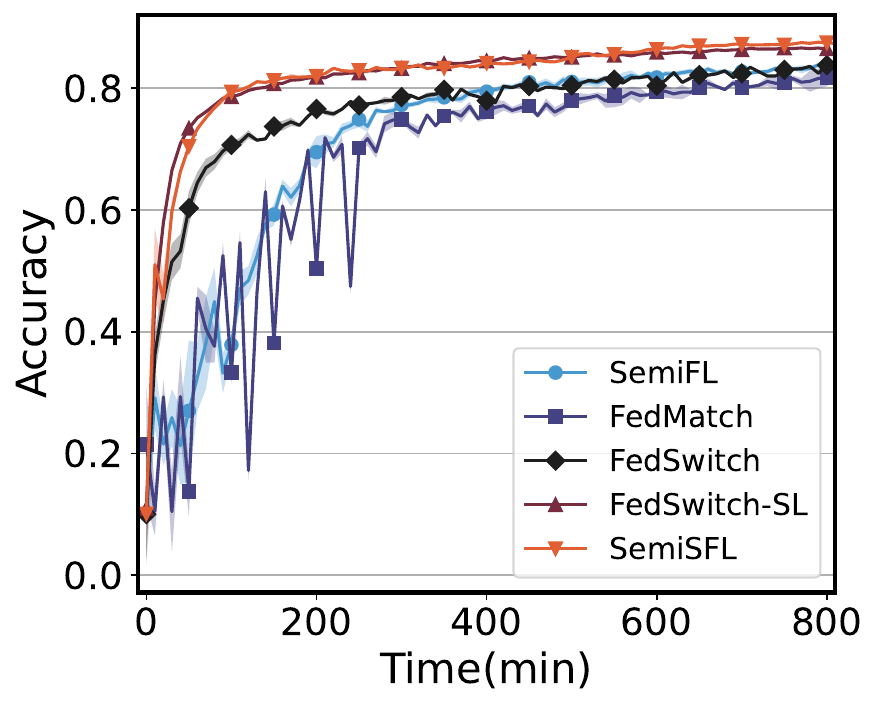}
\end{minipage}
}
\subfloat[Dir(0.5)]{
\begin{minipage}[t]{0.24\linewidth}
    \centering
    \includegraphics[scale=0.3]{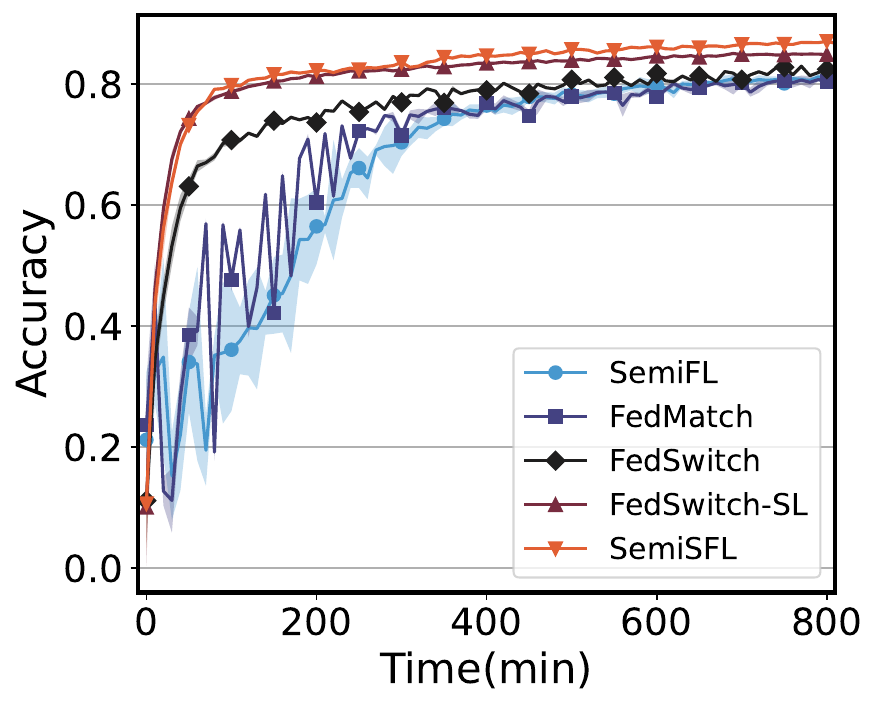}
\end{minipage}
}
\subfloat[Dir(0.1)]{
\begin{minipage}[t]{0.24\linewidth}
    \centering
    \includegraphics[scale=0.3]{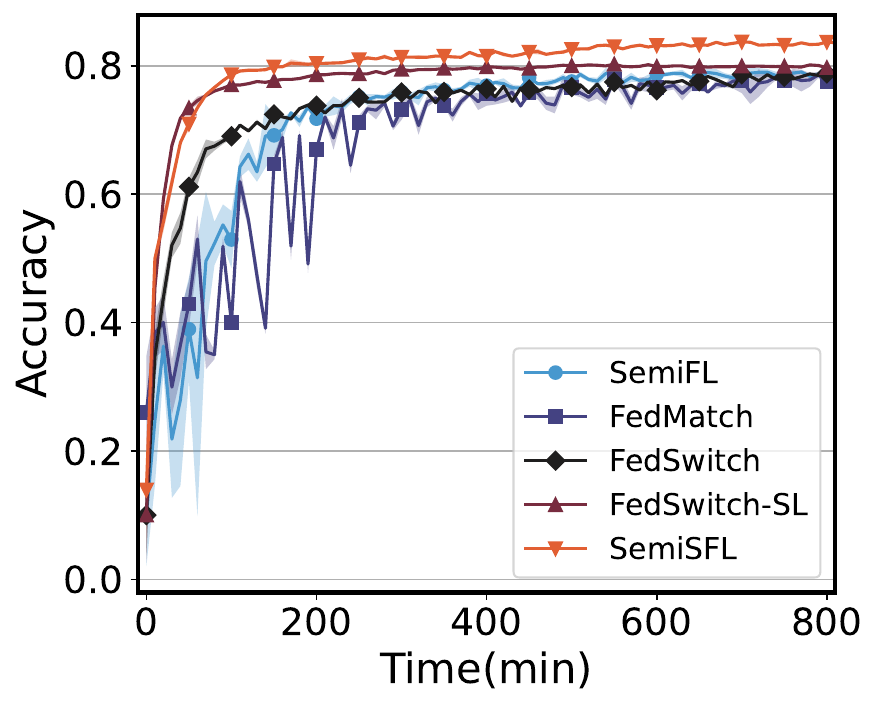}
\end{minipage}
}
\subfloat[Dir(0.05)]{
\begin{minipage}[t]{0.24\linewidth}
    \centering
    \includegraphics[scale=0.3]{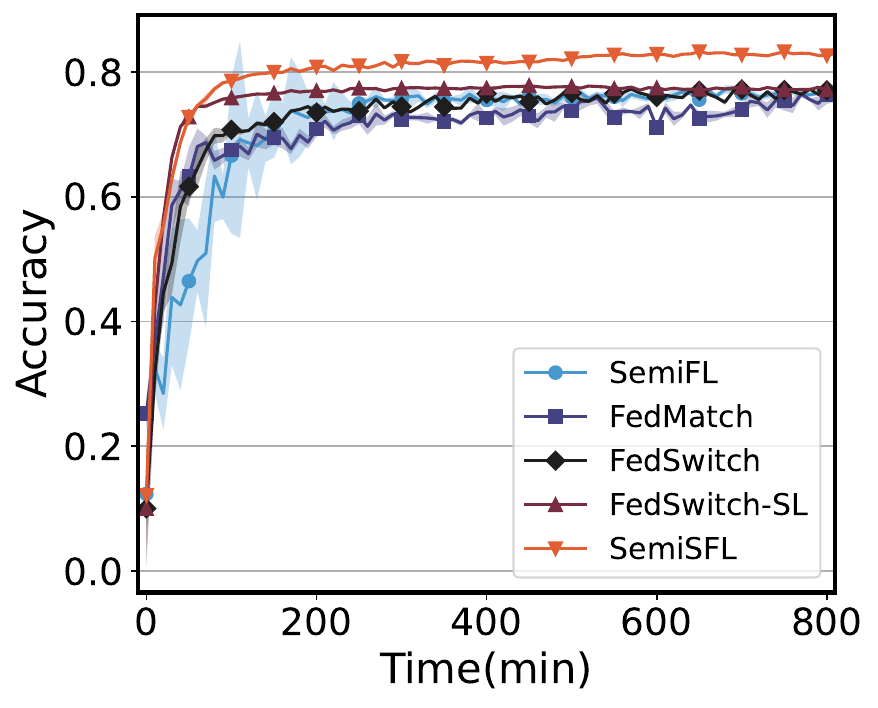}
\end{minipage}
}
\caption{Training process on different data distributions.\label{fig-noniid}}
\vspace{-0.4cm}
\end{figure*}

\subsubsection{Effect of Communication Cost}
We demonstrate the communication efficiency of \method through Fig. \ref{fig-commcost}, which tracks the overall network traffic consumption until the target accuracy is reached. 
Our choice of split layers played a pivotal role in achieving this efficiency. 
For example, in Fig. \ref{fig-commcost}(b), \method consumes only 181.95GB to achieve 80\% accuracy on CIFAR-10, while FedSwitch-FL consumes 59.7GB higher and FL baselines consume from 441.9 to 560.5GB.
This translates to reductions of 24.7\%-83.4\% in communication costs. 
Similarly, Fig. \ref{fig-commcost}(c) shows a 72.9\%-92.9\% reduction in communication costs for \method to reach 85\% accuracy on STL-10, and Fig. \ref{fig-commcost}(d) shows a 73.6\%-83.5\% reduction to reach 60\% accuracy on IMAGE-100.
The low communication cost of both FedSwitch-SL and \method implies the training efficiency of the SFL framework since the communication time is a significant contributor to the overall training time.
However, in certain cases, such as training a customized CNN model, the transmission cost of features might outweigh the benefits of SFL.
Fig. \ref{fig-commcost}(a) shows that \method consumes approximately 74.4GB to achieve 85\% accuracy, which is 6.3 times higher than FedSwitch.

\subsubsection{Adaptability to Data Distribution\label{sec-exp-dir}}

\begin{table}[t]
\centering
\vspace{0.2cm}
\caption{Test accuracy (\%) on different datasets ($\alpha=0.1$).}
\label{tab-noniid}
\renewcommand{\arraystretch}{1.3}
\begin{tabular}{lccc}
\toprule
\multicolumn{1}{c}{\multirow{2}{*}{Baselines}} & \multicolumn{3}{c}{Dataset}                \\ \cline{2-4} 
\multicolumn{1}{c}{}                           & SVHN      & CIFAR-10           & IMAGE-100 \\ \midrule
SemiFL                                         & 80.5(0.7) & 78.9(0.9)          & 50.8(1.7) \\
FedMatch                                       & 84.1(0.6) & 79.7(1.0)          & 29.1(1.7) \\
FedSwitch                                      & 82.9(0.5) & 79.4(0.6)          & 48.9(1.8) \\
FedSwitch-SL                                   & 83.0(0.7) & 79.2(0.7)          & 49.3(1.6) \\
\method                                          & \textbf{87.2(0.2)} & \textbf{83.8(0.2)} & \textbf{56.1(0.4)}  \\ \bottomrule
\end{tabular}
\vspace{-0.2cm}
\end{table}

\begin{table}[t]
\centering
\caption{Test accuracy (\%) of models on CIFAR-10 with different data distributions.}
\label{tab-cifarnoniid}
\renewcommand{\arraystretch}{1.3}
\begin{tabular}{lcccc}
\toprule
\multicolumn{1}{c}{\multirow{2}{*}{Baselines}} & \multicolumn{4}{c}{Data Distribution}         \\ \cline{2-5} 
\multicolumn{1}{c}{}                           & Dir(1.0)  & Dir(0.5)  & Dir(0.1)  & Dir(0.05) \\ \midrule
SemiFL                                         & 84.9(0.9) & 83.1(1.3) & 78.9(0.9) & 76.5(1.1) \\
FedMatch                                       & 85.9(1.0) & 84.4(1.1) & 79.7(1.0) & 76.6(1.4) \\
FedSwitch                                      & 86.2(1.0) & 83.8(1.1) & 79.4(0.6) & 77.3(1.0) \\
FedSwitch-SL                                   & 86.4(1.0) & 84.1(0.7) & 79.2(0.7) & 77.2(1.1) \\
\method                                          & \textbf{87.9(0.2)} & \textbf{86.8(0.2)} & \textbf{83.8(0.2)} & \textbf{82.3(0.1)} \\ \bottomrule
\end{tabular}
\vspace{-0.3cm}
\end{table}

We compare \method with baselines in dealing with data non-IIDness. This is implemented through sampling data from a Dirichlet distribution $Dir(\alpha)$ \cite{hsu2019measuring} for each client.
STL-10 is excluded from this experiment because the necessary ground truth label information for sampling from certain distributions is not available for unlabeled data.
The results are presented in Table \ref{tab-noniid}, where different categories of data on each client follow a distribution of $Dir(0.1)$. 
From the results, \method achieves a consistent improvement over baselines on all three datasets. 
Concretely, \method outperforms FedSwitch-SL by 4.2\%, 4.6\%, and 6.8\% on the three datasets, respectively. 
These results demonstrate the effectiveness of \method in mitigating data non-IIDness, which is getting more significant on datasets with a wider variety of categories.

Moreover, we investigate the adaptability of \method on different data skewness on CIFAR-10. 
The data distributions among clients in our settings are shown in Fig. \ref{fig-hm} and we present the training process on these distributions in Fig. \ref{fig-noniid}. 
Notably, the performance of FedMatch fluctuates dramatically for around 200 min since it uses decomposed model parameters for supervised training and semi-supervised training, highlighting the detrimental effect of such disjoint learning on model convergence.
Table \ref{tab-cifarnoniid} provides a direct comparison of the final performance for different levels of non-IIDness. 
In some extreme cases, FedMatch performs similarly to Supervised-only, suggesting that even with multiple helpers for pseudo-labeling, the model parameters can still be misled by the highly skewed unlabeled dataset. 
A common phenomenon is that the more the data is skewed, the more the model performance deteriorates. 
In contrast, \method consistently achieves the highest accuracy across all levels of data skewness, showcasing its adaptability to diverse data distributions. 
Under extreme non-IID scenarios such as $Dir(0.05)$, the test accuracy of \method improves by 5.0\%-5.8\%, compared to all baselines. 
The experimental results align with our original intention well with the assistance of the clustering of teacher features (features generated from teacher bottom models).
The stable performance over various data distributions is also attributed to the solid foundation built in the \srvUp stage on strongly-augmented data, which is unfortunately often overlooked in current works.
Notably, the results of SVHN and IMAGE-100 are in good agreement with those observed on CIFAR-10. 
Due to the limited space, we omit the experimental results here.


    

\begin{figure}[t]
\centering
\captionsetup{justification=centering}
\subfloat[Test accuracy]{
\begin{minipage}[t]{0.48\linewidth}
    \centering
    \includegraphics[scale=0.28]{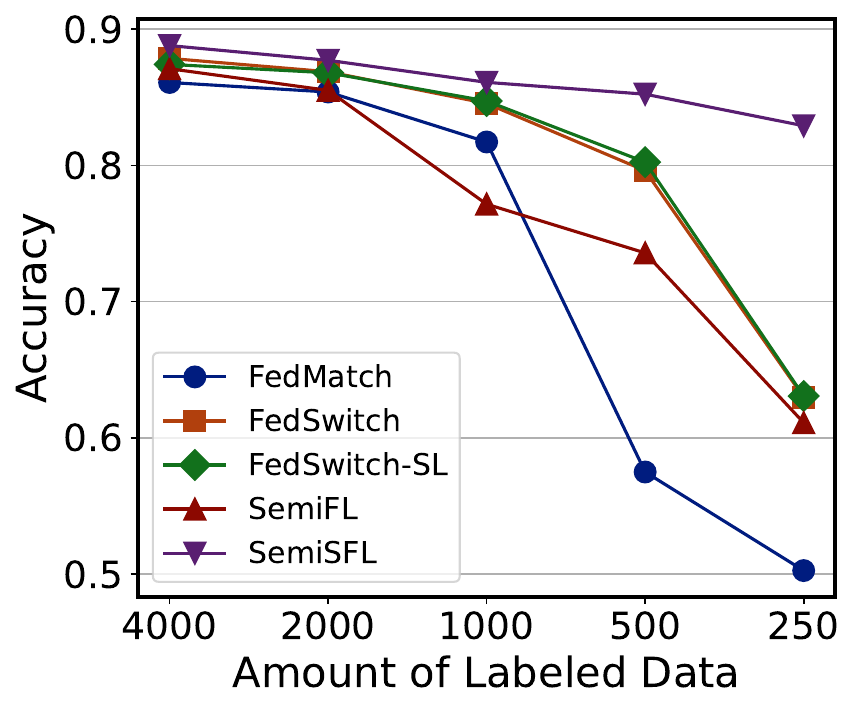}
\end{minipage}
}
\subfloat[Reliability of pseudo-labels]{
\begin{minipage}[t]{0.48\linewidth}
    \centering
    \includegraphics[scale=0.28]{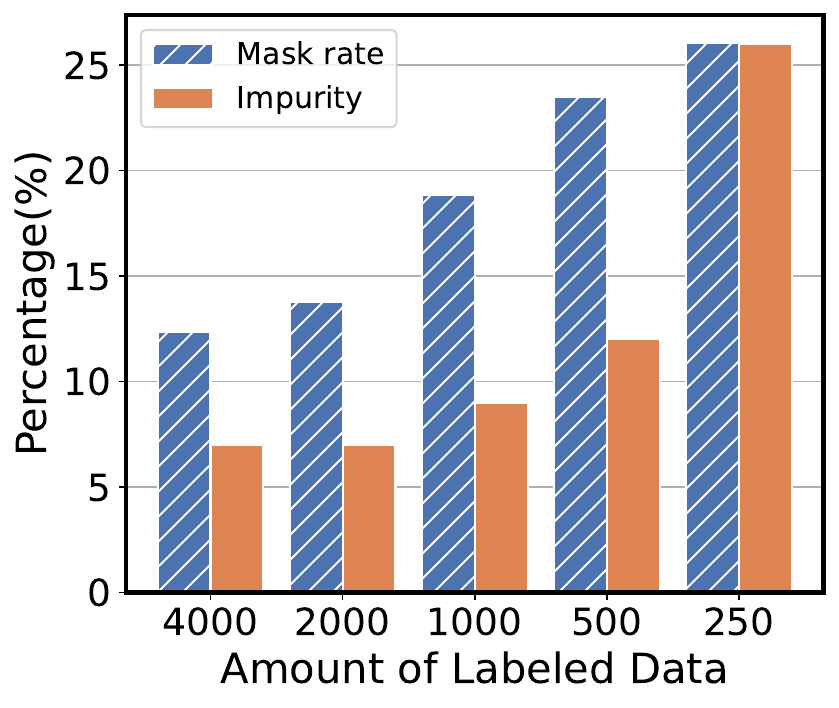}
\end{minipage}
}
\vspace{-0.1cm}
\caption{Training performance varies with data scales.\label{fig-nlabel}}
\vspace{-0.6cm}
\end{figure}

\subsubsection{Impact of the Scale of Labeled Dataset}

We conduct experiments on CIFAR-10, where we place 4,000 labeled data samples on the PS. 
However, in real-world scenarios, the amount of labeled data on the PS varies. 
To explore the impact of different scales of labeled datasets on \method, we conduct experiments by changing the amount of labeled data on the entire dataset. 
The results are presented in Fig. \ref{fig-nlabel}. 
We observe that the test accuracy gradually decreased from 87.9\% to 63.0\% when the amount of labeled data varied from 4,000 to 250 in FedSwitch-SL.
To investigate the reason for the significant performance decline when the amount of labeled data decreased from 500 to 250, we record the mask rate (the number of examples that are masked out) and data impurity (the error rate of unlabeled data that falls above the threshold) of FedSwitch as an example in Fig. \ref{fig-nlabel}(b). 
We infer that this issue is attributed to the increase in both the mask rate and data impurity, causing the data to become unreliable. 

By contrast, \method predominates the quality of predictions, as shown in Fig. \ref{fig-nlabel}(a), indicating the advantages of our proposed clustering regularization. 
\method utilizes data samples with the largest class probability that fall beneath $\tau$, which is considered invalid in consistency regularization. 
Moreover, \srvUp on strong-augmented data provides a solid foundation for semi-supervised learning by enhancing model robustness, thus reducing impurities of unlabeled data. 
In addition, our results suggest that only a few labeled data (less than 2\% of the overall dataset) is enough to satisfy some accuracy requirements, 80\% for example.

\subsection{Ablation Study}
\subsubsection{ Impact of Global Updating Frequency Adaptation}

\begin{figure}[t]
\centering
\captionsetup{justification=centering}
\subfloat[Test accuracy]{
\begin{minipage}[t]{0.48\linewidth}
    \centering
    \includegraphics[scale=0.28]{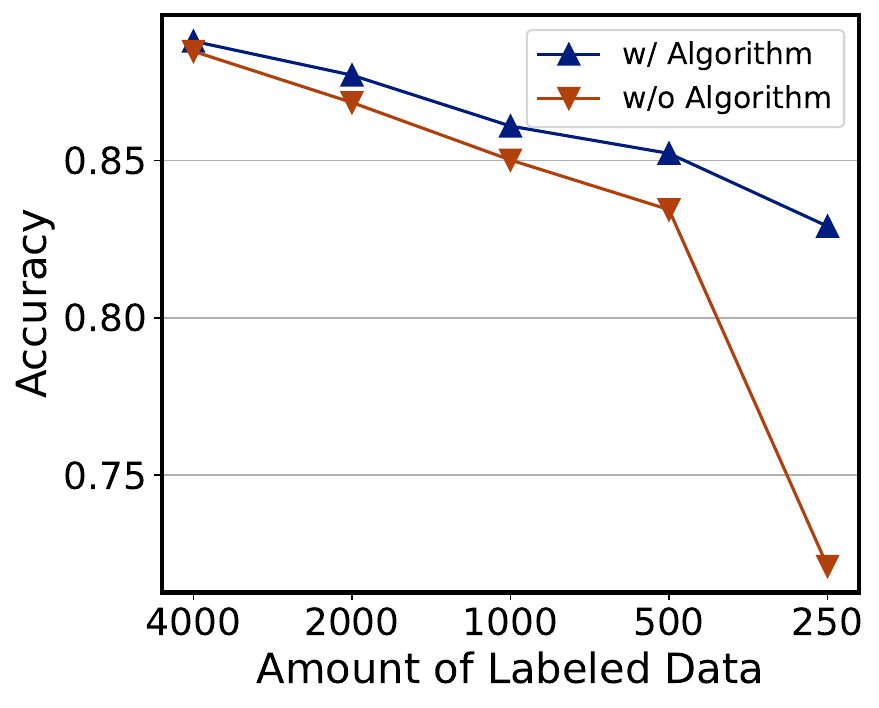}
\end{minipage}
}
\subfloat[$K_s$ distribution]{
\begin{minipage}[t]{0.48\linewidth}
    \centering
    \includegraphics[scale=0.28]{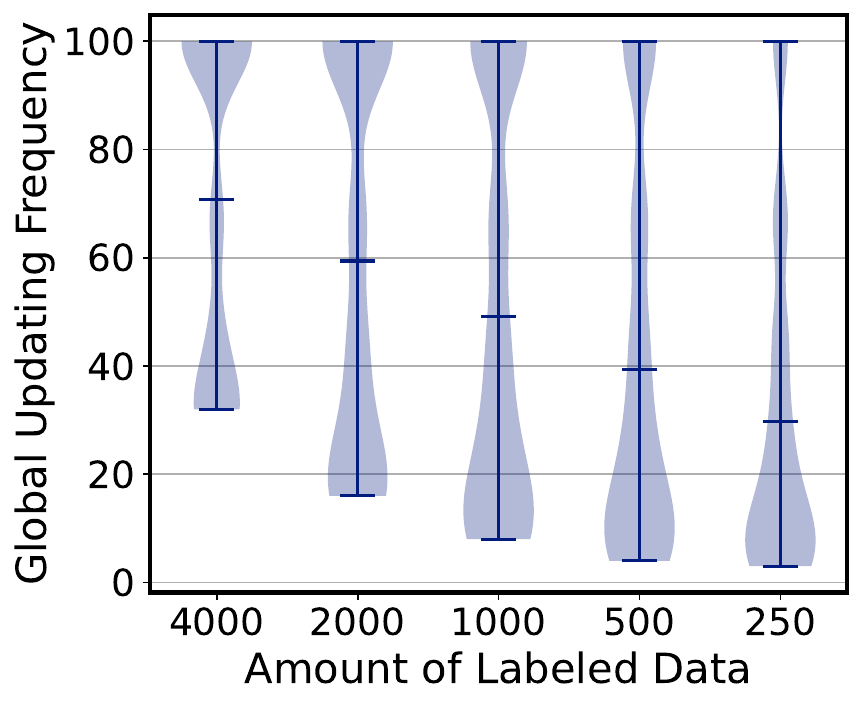}
\end{minipage}
}
\caption{Impact of global updating frequency adaptation.}
\label{fig-algexp}
\vspace{-0.6cm}
\end{figure}

We conduct ablation experiments to assess the impact of our adaptive \InvReg frequency adaptation algorithm on the performance of \method. 
Fig. \ref{fig-algexp} contrasts the test accuracies and distribution of \InvReg frequencies with (w/) and without (w/o) the algorithm. 
As shown in Fig. \ref{fig-algexp}(a), the adaptive algorithm significantly boosts the model’s final accuracy, which validates our analytical results in Section \ref{analysis-section} that the algorithm effectively guides the model towards the global optimum.
The benefits of the algorithm are especially notable in scenarios with scarce labeled data, where rapid saturation of supervised training negatively impacts model performance.
As depicted in Fig. \ref{fig-algexp}(b), in scenarios with fewer labels, the \InvReg frequency tends to stabilize at lower levels. 
For instance, with only 250 labeled samples on PS, the adaptive algorithm improves the accuracy by 10.8\%, with \InvReg frequencies mainly below 30. 
These results demonstrate that the adaptive algorithm substantially enhances training efficiency and model performance in environments with label scarcity by intelligently adjusting updating frequencies. 
Overall, the adaptive algorithm is a vital component in ensuring the training efficiency of \method.

\begin{figure}[t]
\centering
\captionsetup{justification=centering}
\subfloat[Varied EMA decay $\gamma$]{
\begin{minipage}[t]{0.48\linewidth}
    \centering
    \includegraphics[scale=0.28]{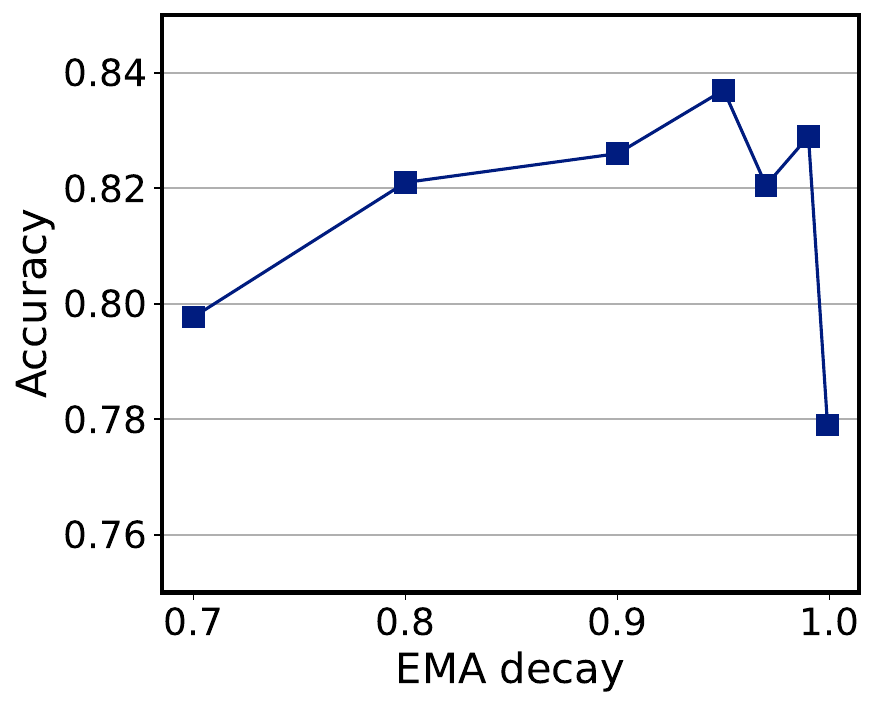}
\end{minipage}
}
\subfloat[Varied confidence
threshold $\tau$]{
\begin{minipage}[t]{0.48\linewidth}
    \centering
    \includegraphics[scale=0.28]{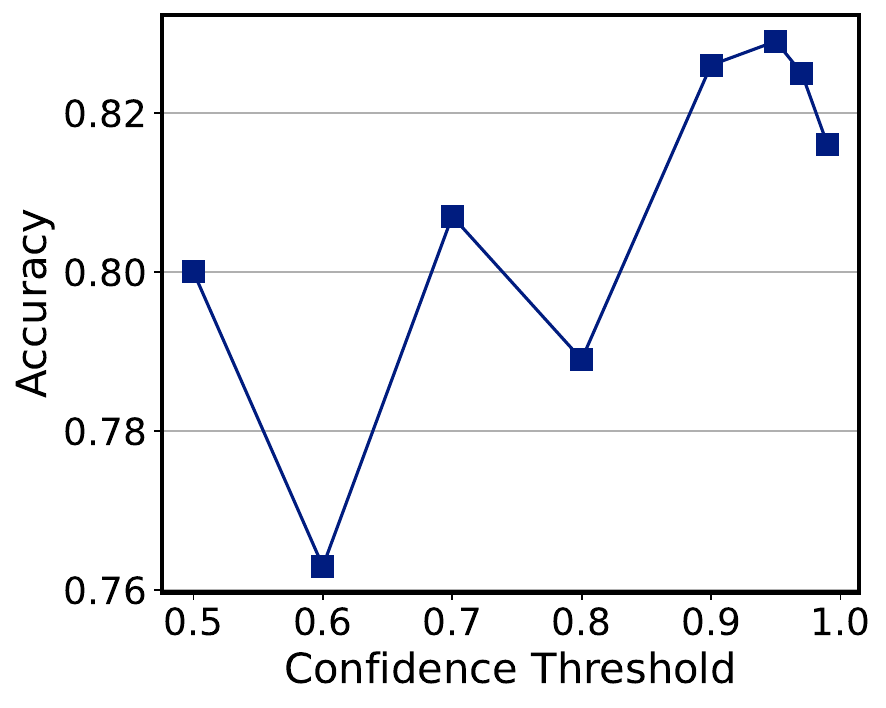}
\end{minipage}
}
\vspace{-0.1cm}
\caption{Plots of ablation studies.}
\label{fig-abs}
\vspace{-0.3cm}
\end{figure}

\begin{table}[t]
\centering
\renewcommand{\arraystretch}{1.3} 
\caption{Test accuracy (\%) under various data distributions with different projection head designs.\label{tab-ph}}
\begin{tabular}{lcccc}
\toprule
\multicolumn{1}{c}{\multirow{2}{*}{Ablation}} & \multicolumn{4}{c}{Data Distribution}                         \\ \cline{2-5} 
\multicolumn{1}{c}{}    & Dir(1.0)      & Dir(0.5)      & Dir(1.0)      & Dir(1.0)      \\ \midrule
No Proj Head            & 86.6          & 82.3          & 82.8          & 80.9          \\
Linear Proj Head        & 87.2          & 86.0          & 82.9          & 81.7          \\
MLP Proj Head           & \textbf{87.9} & \textbf{86.8} & \textbf{83.8} & \textbf{82.3} \\ \bottomrule
\end{tabular}
\vspace{-0.6cm}
\end{table}

\begin{table*}[t]
\centering
\caption{Influence of parameters $\alpha$ and $\beta$ on algorithm performance.\label{tab-algab}}
\resizebox{\textwidth}{!}{
\begin{minipage}{\textwidth}
\centering
\subfloat[SVHN]{
\begin{tabular}{cccc}
\toprule
\textbf{Accuracy(\%)} & $\beta = 4$ & $\beta = 8$ & $\beta = 12$ \\ \midrule
$\alpha = 1.5$ & 91.8 & 91.1 & 91.3 \\
$\alpha = 2.0$ & 91.6 & 91.4 & 91.2 \\
$\alpha = 3.0$ & \textbf{91.8} & 91.0 & 91.4 \\
$\alpha = 4.0$ & 91.4 & 91.6 & 91.1 \\
\bottomrule
\end{tabular}
}
\hfill
\subfloat[CIFAR-10 (4000 labels)]{
\begin{tabular}{cccc}
\toprule
\textbf{Accuracy(\%)} & $\beta = 4$ & $\beta = 8$ & $\beta = 12$ \\ \midrule
$\alpha = 1.5$ & 88.1 & 89.1 & 88.1 \\
$\alpha = 2.0$ & 89.6 & 88.6 & 88.4 \\
$\alpha = 3.0$ & \textbf{89.7} & 88.3 & 88.6 \\
$\alpha = 4.0$ & 88.8 & 88.7 & 87.9 \\
\bottomrule
\end{tabular}
}
\hfill
\subfloat[CIFAR-10 (250 labels)]{
\begin{tabular}{cccc}
\toprule
\textbf{Accuracy(\%)} & $\beta = 4$ & $\beta = 8$ & $\beta = 12$ \\ \midrule
$\alpha = 1.5$ & 81.7 & \textbf{82.9} & 82.6 \\
$\alpha = 2.0$ & 81.3 & 81.9 & 80.7 \\
$\alpha = 3.0$ & 77.8 & 80.3 & 78.1 \\
$\alpha = 4.0$ & 73.6 & 80.0 & 75.3 \\
\bottomrule
\end{tabular}
}
\end{minipage}
}
\vspace{-0.3cm}
\end{table*}

\subsubsection{ Impact of Projection Head}
We investigate the impact of various projection head designs on the effectiveness of clustering regularization in \method under diverse non-IID scenarios. 
Experiments are conducted on the CIFAR-10 dataset, which is adjusted to exhibit varying levels of data skewness as detailed in Section \ref{sec-exp-dir}. 
We test several projection head configurations: one without any projection head (effectively serving as a baseline with just an identity connection), one with a single linear layer, and another featuring a multi-layer perceptron (MLP) with two linear layers separated by a ReLU activation function. 
As illustrated in Table \ref{tab-ph}, performance improves progressively from no projection head, to a single linear layer, and peaks with the MLP projection head. 
These findings affirm that projection heads enhance the efficacy of clustering regularization on addressing data non-IIDness issues by acting as a critical, parameterized component of the regularization process. 
Moreover, the simple MLP projection head strikes an optimal balance by significantly boosting performance without imposing excessive computational demands.

\subsubsection{ Ablation Study on Hyperparameters}
In \method, we set the hyperparameters $\gamma = 0.99$, $\tau=0.95$, $\alpha=1.5$, and $\beta=8$ consistently throughout training. 
We conduct a comprehensive study to assess the impact of these parameters on model performance, especially using the CIFAR-10 dataset configured with 250 labels.
Firstly, we explore the effect of the EMA decay factor $\gamma$, crucial for balancing the influence of historical data on the current teacher model's updates.
Typically, $\gamma$ is set close to 1 to ensure a robust smoothing effect in the model's updates.
As shown in Fig. \ref{fig-abs}(a), settings between 0.9 and 0.99 yield the best performance, while the accuracy declines at 0.999 due to the delay in the integration of new information from unlabeled data since the teacher model is only updated on the PS. 
Therefore, we set $\gamma$ at 0.99, consistent with standard practices and comparative baselines that utilize an EMA approach \cite{tarvainen2017mean}.
Next, we evaluate the confidence threshold $\tau$, pivotal for determining the utilization rate and quality of unlabeled data. 
A higher $\tau$ reduces the number of unlabeled data included in consistency regularization while enhancing the reliability of pseudo labels in forming pseudo clusters, as detailed in Section \label{motivation-section}. 
Fig. \ref{fig-abs}(b) reveals that a $\tau$ setting of 0.95 maximizes test accuracy. 
Although our study maintains a fixed $\tau$, it's worth noting that recent research \cite{zhang2021flexmatch, wang2022freematch} dynamically adjusts $\tau$, which diverges from our focus.
Lastly, we assess the parameters within our global updating frequency adaptation algorithm, $\alpha$ and $\beta$. 
Through systematic testing across datasets such as SVHN and CIFAR-10 with variable label counts, we analyze the impacts of different $\alpha$ and $\beta$ settings. 
Results are summarized in Table \ref{tab-algab}, showing that no single configuration consistently excels.
However, $\alpha=1.5$ and $\beta=8$ are found to be reliably effective, providing a stable and efficient balance for model training.

\section{Related Work}\label{relwork}
\subsection{Federated Learning}
Federated Learning (FL) has been proposed to perform privacy-preserving distributed model training \cite{liao2023adaptive, konevcny2016federated, liu2022enhancing, xu2022adaptive}. 
In FL, data heterogeneity is a crucial factor for training performance, both in terms of accuracy and efficiency. 
Zhao \etal \cite{zhao2018federated} demonstrate that the issue of accuracy reduction caused by non-IID data can be explained by weight divergence, and they address this issue by sharing a small subset of data between clients for local training, which raises concerns about data privacy.
Li \etal \cite{li2020federated} propose adding a proximal term to the objective to improve convergence stability, but it introduces non-negligible computation burdens on clients.
Recently, Li \etal \cite{li2021model} address the non-IID issue from a model representation perspective by correcting local updates with the representation of the global model. 
However, typical FL generally performs model training on labeled data, which is impractical in many real-world applications \cite{lin2014microsoft, radford2019language}.

\subsection{Semi-Supervised Federated Learning}
To utilize unlabeled data in FL, Semi-supervised Federated Learning (Semi-FL) has gained significant attention.
It shares the objective of FL and extends to the semi-supervised learning setting. 
Some studies assume clients possess both labeled and unlabeled data \cite{lin2021semifed, kim2022federated, wang2022enhancing} for their loss calculation.
To relax the assumption of labeled data on clients, several approaches focus on more common scenarios where labeled data reside on the server.
Diao \etal \cite{diao2022semifl} present SemiFL, which applies Mixup technique \cite{berthelot2019mixmatch} to augment the local dataset on clients and pioneer the alternate training phases widely adopted in later literature. 
Moreover, SemiFL achieves accuracy comparable with standalone training. 
Jeong \etal \cite{jeong2021federated} introduce FedMatch, which enforces prediction consistency between clients to address non-IID challenges.
To mitigate the high communication cost in sharing client models, Wang \etal \cite{wang2023knowledge} conduct knowledge distillation on the server, which requires a pre-trained model for the process.
Concurrently, Zhao \etal \cite{zhao2023does} achieve state-of-the-art results by adaptively switching between teacher and student models for pseudo-labeling.
Nonetheless, this method relies on an IIDness hyper-parameter for system control, which might be inaccessible for new tasks or datasets.

\subsection{Split Federated Learning}
Split Federated Learning (SFL) \cite{gupta2018distributed} focuses on training large-scale models on resource-constrained edge devices by incorporating the advantages of federated learning \cite{konevcny2016federated} and split learning. 
For instance, Thapa \etal \cite{thapa2022splitfed} pioneer feasible SFL, which protects model privacy through model splitting and enables client-side model updates through the exchange of features/gradients.
To reduce the traffic consumption in SFL, Han \etal \cite{han2021accelerating} introduce an auxiliary network to update client-side models, without requiring gradients from the server.
Then, Oh \etal \cite {oh2022locfedmix} propose LocFedMix-SL, which uploads mixup-augmented features to the server for faster convergence and employs an auxiliary network to regularize the client-side models for better performance.
To address system heterogeneity, Liao \etal \cite{liao2023accelerating} accelerate SFL on resource-constrained and heterogeneous clients by determining diverse batch sizes for different clients and adapting local updating frequency in each aggregation round.
Recently, Han \etal \cite{han2023splitgp} propose SplitGP to ensure both personalization and generalization capabilities by allocating different tasks for each clients.
Despite these notable advancements, none of the existing SFL works have yet explored utilizing features to address the non-IID issue.

\section{Conclusion}\label{conclu}
In this paper, we have reviewed the distinct properties of SFL and proposed a novel Semi-supervised SFL system, termed \method, to perform model training on unlabeled and non-IID client data by incorporating clustering regularization.
We have theoretically and experimentally investigated the impact of global updating frequency on model convergence.
Then, we have developed a control algorithm for dynamically adjusting the global updating frequency, so as to mitigate the training inconsistency and enhance training performance.
Extensive experiments have demonstrated that \method provides a 3.8× speed-up in training time and reduces the communication cost by about 73.6\% while reaching the target accuracy, and achieves up to 5.8\% improvement in accuracy under non-IID scenarios compared to the baselines.




\balance
\bibliographystyle{IEEEtran}
\bibliography{main}

\begin{thebibliography}{10}
\providecommand{\url}[1]{#1}
\csname url@samestyle\endcsname
\providecommand{\newblock}{\relax}
\providecommand{\bibinfo}[2]{#2}
\providecommand{\BIBentrySTDinterwordspacing}{\spaceskip=0pt\relax}
\providecommand{\BIBentryALTinterwordstretchfactor}{4}
\providecommand{\BIBentryALTinterwordspacing}{\spaceskip=\fontdimen2\font plus
\BIBentryALTinterwordstretchfactor\fontdimen3\font minus
  \fontdimen4\font\relax}
\providecommand{\BIBforeignlanguage}[2]{{%
\expandafter\ifx\csname l@#1\endcsname\relax
\typeout{** WARNING: IEEEtran.bst: No hyphenation pattern has been}%
\typeout{** loaded for the language `#1'. Using the pattern for}%
\typeout{** the default language instead.}%
\else
\language=\csname l@#1\endcsname
\fi
#2}}
\providecommand{\BIBdecl}{\relax}
\BIBdecl

\bibitem{konevcny2016federated}
J.~Kone{\v{c}}n{\`y}, H.~B. McMahan, D.~Ramage, and P.~Richt{\'a}rik,
  ``Federated optimization: Distributed machine learning for on-device
  intelligence,'' \emph{arXiv preprint arXiv:1610.02527}, 2016.

\bibitem{li2019edge}
E.~Li, L.~Zeng, Z.~Zhou, and X.~Chen, ``Edge ai: On-demand accelerating deep
  neural network inference via edge computing,'' \emph{IEEE Transactions on
  Wireless Communications}, vol.~19, no.~1, pp. 447--457, 2019.

\bibitem{xu2022adaptive}
Y.~Xu, Y.~Liao, H.~Xu, Z.~Ma, L.~Wang, and J.~Liu, ``Adaptive control of local
  updating and model compression for efficient federated learning,'' \emph{IEEE
  Transactions on Mobile Computing}, 2022.

\bibitem{liang2022new}
B.~Liang, J.~Cai, and H.~Yang, ``A new cell group clustering algorithm based on
  validation \& correction mechanism,'' \emph{Expert Systems with
  Applications}, vol. 193, p. 116410, 2022.

\bibitem{wen2023survey}
J.~Wen, Z.~Zhang, Y.~Lan, Z.~Cui, J.~Cai, and W.~Zhang, ``A survey on federated
  learning: challenges and applications,'' \emph{International Journal of
  Machine Learning and Cybernetics}, vol.~14, no.~2, pp. 513--535, 2023.

\bibitem{li2020review}
L.~Li, Y.~Fan, M.~Tse, and K.-Y. Lin, ``A review of applications in federated
  learning,'' \emph{Computers \& Industrial Engineering}, vol. 149, p. 106854,
  2020.

\bibitem{liao2023accelerating}
Y.~Liao, Y.~Xu, H.~Xu, Z.~Yao, L.~Wang, and C.~Qiao, ``Accelerating federated
  learning with data and model parallelism in edge computing,'' \emph{IEEE/ACM
  Transactions on Networking}, 2023.

\bibitem{han2021accelerating}
D.-J. Han, H.~I. Bhatti, J.~Lee, and J.~Moon, ``Accelerating federated learning
  with split learning on locally generated losses,'' in \emph{ICML 2021
  Workshop on Federated Learning for User Privacy and Data Confidentiality.
  ICML Board}, 2021.

\bibitem{thapa2022splitfed}
C.~Thapa, P.~C.~M. Arachchige, S.~Camtepe, and L.~Sun, ``Splitfed: When
  federated learning meets split learning,'' in \emph{Proceedings of the AAAI
  Conference on Artificial Intelligence}, vol.~36, no.~8, 2022, pp. 8485--8493.

\bibitem{liao2024mergesfl}
Y.~Liao, Y.~Xu, H.~Xu, L.~Wang, Z.~Yao, and C.~Qiao, ``Mergesfl: Split
  federated learning with feature merging and batch size regulation,'' in
  \emph{2024 IEEE 40th International Conference on Data Engineering
  (ICDE)}.\hskip 1em plus 0.5em minus 0.4em\relax IEEE, 2024, pp. 2054--2067.

\bibitem{simonyan2014very}
K.~Simonyan and A.~Zisserman, ``Very deep convolutional networks for
  large-scale image recognition,'' \emph{arXiv preprint arXiv:1409.1556}, 2014.

\bibitem{zhao2018federated}
Y.~Zhao, M.~Li, L.~Lai, N.~Suda, D.~Civin, and V.~Chandra, ``Federated learning
  with non-iid data,'' \emph{arXiv preprint arXiv:1806.00582}, 2018.

\bibitem{wang2023distribution}
Y.~Wang, Y.~Tong, Z.~Zhou, R.~Zhang, S.~J. Pan, L.~Fan, and Q.~Yang,
  ``Distribution-regularized federated learning on non-iid data,'' in
  \emph{2023 IEEE 39th International Conference on Data Engineering
  (ICDE)}.\hskip 1em plus 0.5em minus 0.4em\relax IEEE, 2023, pp. 2113--2125.

\bibitem{liao2023decentralized}
Y.~Liao, Y.~Xu, H.~Xu, L.~Wang, C.~Qian, and C.~Qiao, ``Decentralized federated
  learning with adaptive configuration for heterogeneous participants,''
  \emph{IEEE Transactions on Mobile Computing}, 2023.

\bibitem{gui2023sk}
J.~Gui, Y.~Song, Z.~Wang, C.~He, and Q.~Huang, ``Sk-gradient: Efficient
  communication for distributed machine learning with data sketch,'' in
  \emph{2023 IEEE 39th International Conference on Data Engineering
  (ICDE)}.\hskip 1em plus 0.5em minus 0.4em\relax IEEE, 2023, pp. 2372--2385.

\bibitem{lin2021semifed}
H.~Lin, J.~Lou, L.~Xiong, and C.~Shahabi, ``Semifed: Semi-supervised federated
  learning with consistency and pseudo-labeling,'' \emph{arXiv preprint
  arXiv:2108.09412}, 2021.

\bibitem{kim2022federated}
W.~Kim, K.~Park, K.~Sohn, R.~Shu, and H.-S. Kim, ``Federated semi-supervised
  learning with prototypical networks,'' \emph{arXiv preprint
  arXiv:2205.13921}, 2022.

\bibitem{wang2022enhancing}
L.~Wang, Y.~Xu, H.~Xu, J.~Liu, Z.~Wang, and L.~Huang, ``Enhancing federated
  learning with in-cloud unlabeled data,'' in \emph{2022 IEEE 38th
  International Conference on Data Engineering (ICDE)}.\hskip 1em plus 0.5em
  minus 0.4em\relax IEEE, 2022, pp. 136--149.

\bibitem{lin2020ensemble}
T.~Lin, L.~Kong, S.~U. Stich, and M.~Jaggi, ``Ensemble distillation for robust
  model fusion in federated learning,'' \emph{Advances in Neural Information
  Processing Systems}, vol.~33, pp. 2351--2363, 2020.

\bibitem{albaseer2020exploiting}
A.~Albaseer, B.~S. Ciftler, M.~Abdallah, and A.~Al-Fuqaha, ``Exploiting
  unlabeled data in smart cities using federated edge learning,'' in \emph{2020
  International Wireless Communications and Mobile Computing (IWCMC)}.\hskip
  1em plus 0.5em minus 0.4em\relax IEEE, 2020, pp. 1666--1671.

\bibitem{zhang2021improving}
Z.~Zhang, Y.~Yang, Z.~Yao, Y.~Yan, J.~E. Gonzalez, K.~Ramchandran, and M.~W.
  Mahoney, ``Improving semi-supervised federated learning by reducing the
  gradient diversity of models,'' in \emph{2021 IEEE International Conference
  on Big Data (Big Data)}.\hskip 1em plus 0.5em minus 0.4em\relax IEEE, 2021,
  pp. 1214--1225.

\bibitem{diao2021semifl}
E.~Diao, J.~Ding, and V.~Tarokh, ``Semifl: Communication efficient
  semi-supervised federated learning with unlabeled clients,'' \emph{arXiv
  preprint arXiv:2106.01432}, 2021.

\bibitem{jeong2021federated}
W.~Jeong, J.~Yoon, E.~Yang, and S.~J. Hwang, ``Federated semi-supervised
  learning with inter-client consistency \& disjoint learning,'' in \emph{9th
  International Conference on Learning Representations, ICLR 2021}.\hskip 1em
  plus 0.5em minus 0.4em\relax International Conference on Learning
  Representations, ICLR, 2021.

\bibitem{long2020fedsiam}
Z.~Long, L.~Che, Y.~Wang, M.~Ye, J.~Luo, J.~Wu, H.~Xiao, and F.~Ma, ``Fedsiam:
  Towards adaptive federated semi-supervised learning,'' \emph{arXiv preprint
  arXiv:2012.03292}, 2020.

\bibitem{zhao2023does}
J.~Zhao, S.~Ghosh, A.~Bharadwaj, and C.-Y. Ma, ``When does the student surpass
  the teacher? federated semi-supervised learning with teacher-student ema,''
  \emph{arXiv preprint arXiv:2301.10114}, 2023.

\bibitem{wang2023knowledge}
J.~Wang, S.~Zeng, Z.~Long, Y.~Wang, H.~Xiao, and F.~Ma, ``Knowledge-enhanced
  semi-supervised federated learning for aggregating heterogeneous lightweight
  clients in iot,'' in \emph{Proceedings of the 2023 SIAM International
  Conference on Data Mining (SDM)}.\hskip 1em plus 0.5em minus 0.4em\relax
  SIAM, 2023, pp. 496--504.

\bibitem{li2021fedbn}
X.~Li, M.~Jiang, X.~Zhang, M.~Kamp, and Q.~Dou, ``Fedbn: Federated learning on
  non-iid features via local batch normalization,'' \emph{arXiv preprint
  arXiv:2102.07623}, 2021.

\bibitem{gupta2018distributed}
O.~Gupta and R.~Raskar, ``Distributed learning of deep neural network over
  multiple agents,'' \emph{Journal of Network and Computer Applications}, vol.
  116, pp. 1--8, 2018.

\bibitem{sohn2020fixmatch}
K.~Sohn, D.~Berthelot, N.~Carlini, Z.~Zhang, H.~Zhang, C.~A. Raffel, E.~D.
  Cubuk, A.~Kurakin, and C.-L. Li, ``Fixmatch: Simplifying semi-supervised
  learning with consistency and confidence,'' \emph{Advances in neural
  information processing systems}, vol.~33, pp. 596--608, 2020.

\bibitem{li2021model}
Q.~Li, B.~He, and D.~Song, ``Model-contrastive federated learning,'' in
  \emph{Proceedings of the IEEE/CVF Conference on Computer Vision and Pattern
  Recognition}, 2021, pp. 10\,713--10\,722.

\bibitem{lee2022contrastive}
D.~Lee, S.~Kim, I.~Kim, Y.~Cheon, M.~Cho, and W.-S. Han, ``Contrastive
  regularization for semi-supervised learning,'' in \emph{Proceedings of the
  IEEE/CVF Conference on Computer Vision and Pattern Recognition}, 2022, pp.
  3911--3920.

\bibitem{khosla2020supervised}
P.~Khosla, P.~Teterwak, C.~Wang, A.~Sarna, Y.~Tian, P.~Isola, A.~Maschinot,
  C.~Liu, and D.~Krishnan, ``Supervised contrastive learning,'' \emph{Advances
  in Neural Information Processing Systems}, vol.~33, pp. 18\,661--18\,673,
  2020.

\bibitem{gupta2022understanding}
K.~Gupta, T.~Ajanthan, A.~v.~d. Hengel, and S.~Gould, ``Understanding and
  improving the role of projection head in self-supervised learning,''
  \emph{arXiv preprint arXiv:2212.11491}, 2022.

\bibitem{cubuk2020randaugment}
E.~D. Cubuk, B.~Zoph, J.~Shlens, and Q.~V. Le, ``Randaugment: Practical
  automated data augmentation with a reduced search space,'' in
  \emph{Proceedings of the IEEE/CVF conference on computer vision and pattern
  recognition workshops}, 2020, pp. 702--703.

\bibitem{li2019convergence}
X.~Li, K.~Huang, W.~Yang, S.~Wang, and Z.~Zhang, ``On the convergence of fedavg
  on non-iid data,'' \emph{arXiv preprint arXiv:1907.02189}, 2019.

\bibitem{yang2021achieving}
H.~Yang, M.~Fang, and J.~Liu, ``Achieving linear speedup with partial worker
  participation in non-iid federated learning,'' \emph{arXiv preprint
  arXiv:2101.11203}, 2021.

\bibitem{haddadpour2019convergence}
F.~Haddadpour and M.~Mahdavi, ``On the convergence of local descent methods in
  federated learning,'' \emph{arXiv preprint arXiv:1910.14425}, 2019.

\bibitem{ajalloeian2020convergence}
A.~Ajalloeian and S.~U. Stich, ``On the convergence of sgd with biased
  gradients,'' \emph{arXiv preprint arXiv:2008.00051}, 2020.

\bibitem{netzer2011reading}
Y.~Netzer, T.~Wang, A.~Coates, A.~Bissacco, B.~Wu, and A.~Y. Ng, ``Reading
  digits in natural images with unsupervised feature learning,'' 2011.

\bibitem{krizhevsky2009learning}
A.~Krizhevsky, G.~Hinton \emph{et~al.}, ``Learning multiple layers of features
  from tiny images,'' 2009.

\bibitem{coates2011analysis}
A.~Coates, A.~Ng, and H.~Lee, ``An analysis of single-layer networks in
  unsupervised feature learning,'' in \emph{Proceedings of the fourteenth
  international conference on artificial intelligence and statistics}.\hskip
  1em plus 0.5em minus 0.4em\relax JMLR Workshop and Conference Proceedings,
  2011, pp. 215--223.

\bibitem{diao2022semifl}
E.~Diao, J.~Ding, and V.~Tarokh, ``Semifl: Semi-supervised federated learning
  for unlabeled clients with alternate training,'' \emph{Advances in Neural
  Information Processing Systems}, vol.~35, pp. 17\,871--17\,884, 2022.

\bibitem{russakovsky2015imagenet}
O.~Russakovsky, J.~Deng, H.~Su, J.~Krause, S.~Satheesh, S.~Ma, Z.~Huang,
  A.~Karpathy, A.~Khosla, M.~Bernstein \emph{et~al.}, ``Imagenet large scale
  visual recognition challenge,'' \emph{International journal of computer
  vision}, vol. 115, pp. 211--252, 2015.

\bibitem{krizhevsky2017imagenet}
A.~Krizhevsky, I.~Sutskever, and G.~E. Hinton, ``Imagenet classification with
  deep convolutional neural networks,'' \emph{Communications of the ACM},
  vol.~60, no.~6, pp. 84--90, 2017.

\bibitem{berthelot2019mixmatch}
D.~Berthelot, N.~Carlini, I.~Goodfellow, N.~Papernot, A.~Oliver, and C.~A.
  Raffel, ``Mixmatch: A holistic approach to semi-supervised learning,''
  \emph{Advances in neural information processing systems}, vol.~32, 2019.

\bibitem{xu2024overcoming}
Y.~Xu, Y.~Liao, L.~Wang, H.~Xu, Z.~Jiang, and W.~Zhang, ``Overcoming noisy
  labels and non-iid data in edge federated learning,'' \emph{IEEE Transactions
  on Mobile Computing}, 2024.

\bibitem{loshchilov2016sgdr}
I.~Loshchilov and F.~Hutter, ``Sgdr: Stochastic gradient descent with warm
  restarts,'' \emph{arXiv preprint arXiv:1608.03983}, 2016.

\bibitem{hsu2019measuring}
T.-M.~H. Hsu, H.~Qi, and M.~Brown, ``Measuring the effects of non-identical
  data distribution for federated visual classification,'' \emph{arXiv preprint
  arXiv:1909.06335}, 2019.

\bibitem{tarvainen2017mean}
A.~Tarvainen and H.~Valpola, ``Mean teachers are better role models:
  Weight-averaged consistency targets improve semi-supervised deep learning
  results,'' \emph{Advances in neural information processing systems}, vol.~30,
  2017.

\bibitem{zhang2021flexmatch}
B.~Zhang, Y.~Wang, W.~Hou, H.~Wu, J.~Wang, M.~Okumura, and T.~Shinozaki,
  ``Flexmatch: Boosting semi-supervised learning with curriculum pseudo
  labeling,'' \emph{Advances in Neural Information Processing Systems},
  vol.~34, pp. 18\,408--18\,419, 2021.

\bibitem{wang2022freematch}
Y.~Wang, H.~Chen, Q.~Heng, W.~Hou, Y.~Fan, Z.~Wu, J.~Wang, M.~Savvides,
  T.~Shinozaki, B.~Raj \emph{et~al.}, ``Freematch: Self-adaptive thresholding
  for semi-supervised learning,'' \emph{arXiv preprint arXiv:2205.07246}, 2022.

\bibitem{liao2023adaptive}
Y.~Liao, Y.~Xu, H.~Xu, L.~Wang, and C.~Qian, ``Adaptive configuration for
  heterogeneous participants in decentralized federated learning,'' in
  \emph{IEEE INFOCOM 2023-IEEE Conference on Computer Communications}.\hskip
  1em plus 0.5em minus 0.4em\relax IEEE, 2023, pp. 1--10.

\bibitem{liu2022enhancing}
J.~Liu, Y.~Xu, H.~Xu, Y.~Liao, Z.~Wang, and H.~Huang, ``Enhancing federated
  learning with intelligent model migration in heterogeneous edge computing,''
  in \emph{2022 IEEE 38th International Conference on Data Engineering
  (ICDE)}.\hskip 1em plus 0.5em minus 0.4em\relax IEEE, 2022, pp. 1586--1597.

\bibitem{li2020federated}
T.~Li, A.~K. Sahu, M.~Zaheer, M.~Sanjabi, A.~Talwalkar, and V.~Smith,
  ``Federated optimization in heterogeneous networks,'' \emph{Proceedings of
  Machine learning and systems}, vol.~2, pp. 429--450, 2020.

\bibitem{lin2014microsoft}
T.-Y. Lin, M.~Maire, S.~Belongie, J.~Hays, P.~Perona, D.~Ramanan,
  P.~Doll{\'a}r, and C.~L. Zitnick, ``Microsoft coco: Common objects in
  context,'' in \emph{Computer Vision--ECCV 2014: 13th European Conference,
  Zurich, Switzerland, September 6-12, 2014, Proceedings, Part V 13}.\hskip 1em
  plus 0.5em minus 0.4em\relax Springer, 2014, pp. 740--755.

\bibitem{radford2019language}
A.~Radford, J.~Wu, R.~Child, D.~Luan, D.~Amodei, I.~Sutskever \emph{et~al.},
  ``Language models are unsupervised multitask learners,'' \emph{OpenAI blog},
  vol.~1, no.~8, p.~9, 2019.

\bibitem{oh2022locfedmix}
S.~Oh, J.~Park, P.~Vepakomma, S.~Baek, R.~Raskar, M.~Bennis, and S.-L. Kim,
  ``Locfedmix-sl: Localize, federate, and mix for improved scalability,
  convergence, and latency in split learning,'' in \emph{Proceedings of the ACM
  Web Conference 2022}, 2022, pp. 3347--3357.

\bibitem{han2023splitgp}
D.-J. Han, D.-Y. Kim, M.~Choi, C.~G. Brinton, and J.~Moon, ``Splitgp: Achieving
  both generalization and personalization in federated learning,'' in
  \emph{IEEE INFOCOM 2023-IEEE Conference on Computer Communications}.\hskip
  1em plus 0.5em minus 0.4em\relax IEEE, 2023, pp. 1--10.

\end{thebibliography}

\begin{appendices}
\label{appendix}
\setcounter{section}{0}
\setcounter{lemma}{0}
\setcounter{theorem}{0}
\setcounter{equation}{0}
\section{}
\subsection{Proof of Theorem \ref{theorem1}}
\setcounter{section}{0}
\setcounter{lemma}{0}
\setcounter{theorem}{0}
\setcounter{equation}{0}
\begin{theorem} 
\label{apx-theorem1}
The sequence of outputs $\{\boldsymbol{w}^{h, k}\}$ generated by supervised training and global aggregation satisfies:
\begin{equation}
    \underset{h \in [H], k \in \{0, \cdots, K_s\}}{\min} {\mathbb{E} \Vert \nabla F(\boldsymbol{w}^{h, k}) \Vert^2} 
    \leq \frac{2(F(\boldsymbol{w}^0) - F(\boldsymbol{w}^*))}{\eta_1 H K_s} 
    +\Phi.
    \nonumber
\end{equation}
where $\Phi \triangleq (\frac{L^2(K_u-1)(2K_u-1)\eta_1^2}{3K_s} + \frac{LK_u^2 \eta_1^3}{K_s} + 1) G_u^2 + (L \eta_1 + \frac{2 K_u}{K_s}) G_s^2$.
\end{theorem}

\emph{Proof:} 
First, for the model trained on the supervised stage, according to Lipschitz smoothness property in Assumption \ref{assump1}, we have:
\begin{equation}
\begin{split}
    & \mathbb{E} [F(\boldsymbol{w}^{h, k+1})] - \mathbb{E} [ F(\boldsymbol{w}^{h , k})]  \\
    & \leq -\eta_h \mathbb{E} \langle \nabla F(\boldsymbol{w}^{h, k}),  \tilde{\nabla} f_{s}(\boldsymbol{w}^{h, k}) \rangle 
     + \frac{L \eta_h^2}{2} \mathbb{E} [\Vert \tilde{\nabla} f_{s}(\boldsymbol{w}^{h, k}) \Vert^2]
    \nonumber
\end{split}
\end{equation} 
By using Lemma \ref{apx-lemma5} and summing up $K_s$ global iterations, we obtain:
\begin{equation}
\begin{split}
    & \mathbb{E} [F(\srvmodh)] - \mathbb{E} [ F(\boldsymbol{w}^{h})] \\
    & \leq 
     - \frac{1}{2} \eta_h \sum \limits_{k=0}^{K_s-1} \mathbb{E} \Vert \nabla F(\boldsymbol{w}^{h, k}) \Vert^2
    + \frac{K_s(L \eta_h^2 G_s^2 + \eta_h G_u^2)}{2}
    \nonumber
\end{split}
\end{equation}
Similarly, for the model after aggregation, we have:
\begin{equation}
\begin{split}
    & \mathbb{E} [F(\boldsymbol{w}^{h+1})] - \mathbb{E} [ F(\srvmodh)] \\
    & \leq - \mathbb{E} \langle \nabla F(\srvmodh),  \Delta^{h} \rangle 
     + \frac{L}{2} \mathbb{E} [\Vert \Delta^h \Vert^2] \\
    & \underset{(a)}{\leq} -\eta_h \frac{K_u}{2} \mathbb{E} \Vert \nabla F(\srvmodh) \Vert^2 + \frac{(K_u-1)(2K_u-1)L^2 \eta_h^3 G_u^2}{6} \\
    & + \eta_h K_u G_s^2 + \frac{L}{2} \mathbb{E} [\Vert \Delta^{h} \Vert^2] \\
    \nonumber
\end{split}
\end{equation}
where (a) follows Lemma \ref{apx-lemma5}.

Assume that $\eta_h$ decreases monotonically.
By summing up all aggregation rounds, and from $F(\boldsymbol{w}^H) \leq \mathbb{E} [F(\boldsymbol{w}^*)]$, we can write:
\begin{equation}
\begin{split}
    & \frac{1}{H}\sum\limits_{h=1}^H{\mathbb{E} [K_u\Vert \nabla F(\srvmodh) \Vert^2 + \sum\limits_{k=0}^{K_s-1} \Vert \nabla F(\boldsymbol{w}^{h, k}) \Vert^2]} \\
    & \leq \frac{2(F(\boldsymbol{w}^0) - F(\boldsymbol{w}^*))}{\eta_1 H} \\ 
    & + (\frac{L^2(K_u-1)(2K_u-1)\eta_1^2}{3} + K_s + L K_u^2 \eta_1) G_u^2 \\
    & + (L K_s \eta_1 + 2K_u)G_s^2 \\
    \nonumber
 \end{split}   
\end{equation}
Since $K_u > 0$, we have:
\begin{equation}
\begin{split}
    & \underset{h \in [H], k \in \{0,\cdots, K_s\}}{\min} \mathbb{E} \Vert \nabla F(\boldsymbol{w}^{h, k}) \Vert^2 \\
    & \leq \frac{1}{H K_s}\sum\limits_{k=1}^K\sum\limits_{k=0}^{K_s}{\mathbb{E} \Vert \nabla F(\boldsymbol{w}^{h, k}) \Vert^2} \\
    & \leq \frac{2(F(\boldsymbol{w}^0) - F(\boldsymbol{w}^*))}{\eta_1 H K_s} + \Phi
    \nonumber
\end{split}   
\end{equation}
which completes the proof.

\subsection{Key Lemmas}
\begin{lemma}
For random variables $X_1, \ldots, X_r$, we have:
\label{apx-lemma1}
    \begin{equation}
    \mathbb{E}[\Vert X_1 + \ldots + X_r \Vert^2] \leq r\mathbb{E}[\Vert X_1\Vert^2 + \ldots + \Vert X_r\Vert^2].
    \nonumber
    \end{equation}
\end{lemma}

\begin{lemma}
For random vectors $X, Y, Z$, we have:
\label{apx-lemma2}
    \begin{equation}
    \mathbb{E} \langle X, Y+Z \rangle \geq \mathbb{E} \langle X, Y \rangle - \Vert \mathbb{E} \langle X, Z \rangle \Vert.
    \nonumber
   \end{equation} 
\end{lemma}

\begin{lemma}
\label{apx-lemma3}
    According to Assumption \ref{assump2}, we have:
    \begin{equation}
    \label{apx-lemma3-eq1}
        \mathbb{E} \Vert \tilde{\nabla} f_{u, i}(\boldsymbol{w}) \Vert^2 \leq G_u^2; 
    \end{equation}
    \begin{equation}
    \label{apx-lemma3-eq2}
        \Vert \nabla f_{u}(\boldsymbol{w}) \Vert^2 \leq G_u^2; 
    \end{equation}
    \begin{equation}
    \label{apx-lemma3-eq3}
        \Vert \nabla f_{s}(\boldsymbol{w}) \Vert^2 \leq G_s^2.
    \end{equation}
\end{lemma}

\emph{Proof:} 
For Eq. (\ref{apx-lemma3-eq1}):
\begin{equation}
\begin{split}
    & \mathbb{E} \Vert \nabla f_{u, i}(\boldsymbol{w}) \Vert^2  \\
    & \underset{(a)}{\leq} \frac{1}{|\mathcal{B}_{U, i}|}\sum_{x \in [\mathcal{B}_{U, i}]}{ \Vert \tilde{\nabla} f_{u, i}(\boldsymbol{w}) \Vert^2} \\
    & \underset{(b)}{\leq} G_u^2
    \nonumber
\end{split}
\end{equation}
where (a) is obtained by Lemma \ref{apx-lemma1}, (b) follows Assumption \ref{assump2}. Eq. (\ref{apx-lemma3-eq2}) and Eq. (\ref{apx-lemma3-eq3}) can be proved in a similar way.

\begin{lemma}
\label{apx-lemma4}
    According to Assumptions \ref{assump1} and \ref{assump2}, we have:
    \begin{equation}
    \begin{split}
    \label{apx-lemma2-eq1}
        \mathbb{E} \langle \nabla f_u(\srvmodh), \Delta^{h} \rangle \geq \frac{\eta_h K_u}{2} \mathbb{E} \Vert \nabla f_u(\srvmodh) \Vert^2 \\
        - \frac{(K_u-1)(2K_u-1)L^2 \eta_h^3 G_u^2}{6}.
        \nonumber
    \end{split}
    \end{equation}
    where $\Delta^{h} \triangleq \frac{1}{N}\sum \limits_{i=1}^N {(\boldsymbol{w}_i^{h+1} - \srvmodh)}$.
\end{lemma}

\emph{Proof:} 
\begin{equation}
\begin{split}
    & \mathbb{E} \langle \nabla f_u(\srvmodh), \Delta^{h} \rangle \\
    & = \mathbb{E} \langle \nabla f_u(\srvmodh), \Delta^{h} - \eta_h K_u \nabla f_u(\srvmodh) + \eta_h K_u \nabla f_u(\srvmodh) \rangle \\
    & \underset{(a)}{\geq} \eta K_u \mathbb{E} \Vert \nabla f_u(\srvmodh) \Vert^2 \\ 
    & \quad - \Vert \mathbb{E} \langle \nabla f_u(\srvmodh), \Delta^{h} - \eta_h K_u \nabla f_u(\srvmodh) \rangle \Vert \\
    & \geq \eta_h K_u \mathbb{E} \Vert \nabla f_u(\srvmodh) \Vert^2 \\
    & \quad - \Vert \mathbb{E} \langle \sqrt{\eta_h K_u}\nabla f_u(\srvmodh), \frac{\sqrt{\eta_h}}{\sqrt{K_u}} (\frac{\Delta^{h}}{\eta} - K_u \nabla f_u(\srvmodh)) \rangle \Vert \\
    & \geq \eta_h K_u \mathbb{E} \Vert \nabla f_u(\srvmodh) \Vert^2 \\
    & \quad - \frac{1}{2}(\eta_h K_u \mathbb{E} \Vert \nabla f_u(\srvmodh) \Vert^2 + \frac{\eta_h}{K_u}\mathbb{E} \Vert \frac{\Delta^{h}}{\eta_h} - K_u \nabla f_u(\srvmodh) \Vert^2)
    \label{pf1-lemma4}
\end{split}
\end{equation}
where (a) follows \ref{apx-lemma2}. 
Note that:
\begin{equation}
\begin{split}
    \Delta^{h}
    & = \frac{1}{N}\sum \limits_{i=1}^N\sum \limits_{k=0}^{K_u -1} {(\boldsymbol{w}_i^{h+, k+1} - \boldsymbol{w}_i^{h+, k})}  \\
    & = \frac{1}{N}\sum \limits_{i=1}^N\sum \limits_{k=0}^{K_u-1} {\eta_h \tilde{\nabla} f_{u, i}(\boldsymbol{w}_i^{h+, k})}
    \nonumber
\end{split}
\end{equation}
We have:
\begin{equation}
\begin{split}
    & \frac{\Delta^{h}}{\eta_h} - K_u \nabla f_u(\srvmodh) \\
    & = \frac{1}{N}\sum \limits_{i=1}^N \sum\limits_{k=0}^{K_u-1}{\tilde{\nabla} f_{u, i}(\boldsymbol{w}_i^{h+, k})} - K_u \frac{1}{N}\sum\limits_{i=1}^N {\nabla f_{u,i}(\srvmodh)} \\
    & = \frac{1}{N}\sum \limits_{i=1}^N \sum\limits_{k=0}^{K_u-1}{(\tilde{\nabla} f_{u, i}(\boldsymbol{w}_i^{h+, k}) - \nabla f_{u,i}(\srvmodh)}) \\
    & = \frac{1}{N}\sum \limits_{i=1}^N \sum\limits_{k=0}^{K_u-1}{(\nabla f_{u, i}(\boldsymbol{w}_i^{h+, k}) - \nabla f_{u,i}(\srvmodh)}) \\
    \nonumber
\end{split}
\end{equation}
where the last equality follows Assumption \ref{assump3}. So we can bound $\mathbb{E} \Vert \frac{\Delta^{h}}{\eta_h} - K_u \nabla F(\srvmodh) \Vert^2$ as:
\begin{equation}
\begin{split}
    & \mathbb{E} \Vert \frac{\Delta^{h}}{\eta_h} - K_u \nabla F(\srvmodh) \Vert^2  \\ 
    & \underset{(a)}{\leq} \frac{L^2}{N}\sum \limits_{i=1}^N \sum\limits_{k=0}^{K-1}{\mathbb{E} \Vert \boldsymbol{w}_i^{h+, k} - \boldsymbol{w}_i^{h+} \Vert^2} \\
    & = \frac{L^2}{N}\sum \limits_{i=1}^N \sum\limits_{k=0}^{K-1}{\mathbb{E} \Vert \sum\limits_{m=0}^{k-2}\eta_h \tilde{\nabla} f_{u, i}(\boldsymbol{w}_i^{h+,m}) \Vert^2} \\
    & \underset{(b)}{\leq} \frac{L^2\eta_h^2}{N}\sum \limits_{i=1}^N \sum\limits_{k=0}^{K-1}{ ((k-1)\sum\limits_{m=0}^{k-2} \mathbb{E} \Vert \tilde{\nabla} f_{u, i}(\boldsymbol{w}_i^{h+,m}) \Vert^2)} \\
    & \underset{(c)}{\leq} \frac{L^2}{N}\sum \limits_{i=1}^N \sum\limits_{k=1}^{K_u}{((k-1)^2 \eta_h^2 G_u^2)} \\
    & = \frac{K_u(K_u-1)(2K_u-1)L^2 \eta_h^2 G_u^2}{6}
    \label{pf2-lemma4}
\end{split}
\end{equation}
where (a) and (b) are obtained by Lemma \ref{apx-lemma1} and follows Assumption \ref{assump1}, (c) follows Lemma \ref{apx-lemma3}.

\begin{lemma}
\label{apx-lemma5}
    According to Assumptions \ref{assump1} and \ref{assump2}, we have:
    \begin{equation}
    \label{apx-lemma5-eq1}
        \mathbb{E} \langle \nabla F(\boldsymbol{w}^{h}),  \tilde{\nabla} f_{s}(\boldsymbol{w}^{h}) \rangle \geq \frac{1}{2} \mathbb{E} \Vert \nabla F(\boldsymbol{w}^{h}) \Vert^2 - \frac{1}{2} G_u^2;
    \end{equation}
    \begin{equation}
    \label{apx-lemma5-eq2}
    \begin{split}
        \mathbb{E} \langle \nabla F(\srvmodh)&, \Delta^h \rangle 
         \geq \frac{\eta_h K_u}{2} \mathbb{E} \Vert \nabla F(\srvmodh) \Vert^2 \\
        & - \frac{(K_u-1)(2K_u-1)L^2 \eta_h^3 G_u^2}{6} -\eta_h K_u G_s^2.
    \end{split}
    \end{equation}
    where $\Delta^{h} \triangleq \frac{1}{N}\sum \limits_{i=1}^N {(\boldsymbol{w}_i^{h + 1} - \srvmodh)}$.
\end{lemma}

\emph{Proof:} 
For Eq. (\ref{apx-lemma5-eq1}), we have:
\begin{equation}
\begin{split}
    & \mathbb{E} \langle \nabla F(\boldsymbol{w}^{h}),  \tilde{\nabla} f_{s}(\boldsymbol{w}^{h}) \rangle \\
    & \underset{(a)}{=} \mathbb{E} \langle \nabla F(\boldsymbol{w}^{h}),  \nabla f_{s}(\boldsymbol{w}^{h}) \rangle \\
    & = \mathbb{E} \langle \nabla F(\boldsymbol{w}^{h}), -\nabla f_u(\boldsymbol{w}^{h}) + \nabla F(\boldsymbol{w}^{h}) \rangle \\
    & \underset{(b)}{\geq} \mathbb{E} \Vert \nabla F(\boldsymbol{w}^{h}) \Vert^2 - \Vert \mathbb{E} \langle \nabla F(\boldsymbol{w}^{h}), -\nabla f_u(\boldsymbol{w}^h) \rangle \Vert \\
    \nonumber
\end{split}
\end{equation}
where (a) follows Assumption \ref{assump3}, (b) comes from Lemma \ref{apx-lemma2}.

Note that $2 \langle U, V \rangle \leq \Vert U \Vert^2 +\Vert V \Vert^2$, we obtain:
\begin{equation}
\begin{split}
    & \Vert \mathbb{E} \langle \nabla F(\boldsymbol{w}^{h}), -\nabla f_u(\boldsymbol{w}^{h}) \rangle \Vert \\
    & \leq \frac{1}{2} (\mathbb{E} \Vert \nabla F(\boldsymbol{w}^{h}) \Vert^2 + \mathbb{E} \Vert \nabla f_u(\boldsymbol{w}^{h}) \Vert^2) \\
    & \leq \frac{1}{2} \mathbb{E} \Vert \nabla F(\boldsymbol{w}^{h}) \Vert^2 + \frac{1}{2} G_u^2
    \nonumber
\end{split}
\end{equation}
where the last inequality follows Lemma \ref{apx-lemma1}.
According to the above inequalities, we have:
\begin{equation}
    \mathbb{E} \langle \nabla F(\boldsymbol{w}^{h}),  \tilde{\nabla} f_{s}(\boldsymbol{w}^{h}) \rangle \geq \frac{1}{2} \mathbb{E} \Vert \nabla F(\boldsymbol{w}^{h}) \Vert^2 - \frac{1}{2} G_u^2
    \nonumber
\end{equation}
which completes the proof of Eq. (\ref{apx-lemma5-eq1}). 

For Eq. (\ref{apx-lemma5-eq2}), similar to Eq. (\ref{pf1-lemma4}), we have:
\begin{equation}
\begin{split}
    & \mathbb{E} \langle \nabla F(\srvmodh), \Delta^{h} \rangle \\
    & \geq \eta_h K_u \mathbb{E} \Vert \nabla F(\srvmodh) \Vert^2 \\
    & \quad - \frac{1}{2}(\eta_h K_u \mathbb{E} \Vert \nabla F(\srvmodh) \Vert^2 + \frac{\eta_h}{K_u}\mathbb{E} \Vert \frac{\Delta^{h}}{\eta_h} - K_u \nabla F(\srvmodh) \Vert^2)
    \nonumber
\end{split}
\end{equation}
By letting:
\begin{equation}
\begin{split}
    & X = \frac{1}{N}\sum \limits_{i=1}^N \sum\limits_{k=0}^{K-1}{(\tilde{\nabla} f_{u, i}(\boldsymbol{w}_i^{h+, k}) - \nabla f_u(\srvmodh)}) \\
    & Y = \frac{1}{N} \sum\limits_{i=1}^N\sum\limits_{k=0}^{K-1} (\nabla f_u(\srvmodh) - \nabla F(\srvmodh)) \\
    \nonumber
\end{split}
\end{equation}
we have:
\begin{equation}
\begin{split}
    \frac{\Delta^{h}}{\eta_h} - K_u \nabla F(\srvmodh) = X + Y
    \nonumber
\end{split}
\end{equation}
So we bound $\mathbb{E} \Vert \frac{\Delta^{h}}{\eta_h} - K_u \nabla F(\srvmodh) \Vert^2$ as:
\begin{equation}
\begin{split}
    & \mathbb{E} \Vert \frac{\Delta^{h}}{\eta_h} - K_u \nabla F(\srvmodh) \Vert^2  \\
    & \leq 2 \mathbb{E} \Vert X \Vert^2 + 2 \mathbb{E} \Vert Y \Vert^2 \\
    & \leq 2 \mathbb{E} \Vert X \Vert^2 + 2 K_u^2 G_s^2
    \nonumber
\end{split}
\end{equation}
where the above inequalities are obtained by Lemma \ref{apx-lemma1} and Lemma \ref{apx-lemma3}. For $\mathbb{E} \Vert X \Vert^2$, it can be bounded as Eq. (\ref{pf2-lemma4}).

Putting the pieces together, we get:
\begin{equation}
\begin{split}
    \mathbb{E} \langle \nabla F(\srvmodh), & \Delta^h \rangle
     \geq \frac{\eta_h K_u}{2}\mathbb{E} \Vert \nabla F(\srvmodh) \Vert^2 \\
    & - \frac{(K_u-1)(2K_u-1)L^2 \eta_h^3 G_u^2}{6} - \eta_h K_u G_s^2
    \nonumber
\end{split}
\end{equation}
which completes the proof.
\end{appendices}

\end{document}